# Network representation learning systematic review: ancestors and current development state

Amina Amara[1, a], Mohamed Ali Hadj Taieb[1, b], Mohamed Ben Aouicha[1, c]

[1] Data Engineering and Semantics Research Unit, Faculty of Sciences of Sfax, University of Sfax, Tunisia

[a] amara.amina@hotmail.com, [b] mohamedali.hajtaieb@fss.usf.tn, [c] mohamed.benaouicha@fss.usf.tn

## Abstract

Real-world information networks are increasingly occurring across various disciplines including online social networks and citation networks. These network data are generally characterized by sparseness, nonlinearity and heterogeneity bringing different challenges to the network analytics task to capture inherent properties from network data. Artificial intelligence and machine learning have been recently leveraged as powerful systems to learn insights from network data and deal with presented challenges. As part of machine learning techniques, graph embedding approaches are originally conceived for graphs constructed from feature represented datasets, like image dataset, in which links between nodes are explicitly defined. These traditional approaches cannot cope with network data challenges. As a new learning paradigm, network representation learning has been proposed to map a real-world information network into a low-dimensional space while preserving inherent properties of the network. In this paper, we present a systematic comprehensive survey of network representation learning, known also as network embedding, from birth to the current development state. Through the undertaken survey, we provide a comprehensive view of reasons behind the emergence of network embedding and, types of settings and models used in the network embedding pipeline. Thus, we introduce a brief history of representation learning and word representation learning ancestor of network embedding. We provide also formal definitions of basic concepts required to understand network representation learning followed by a description of network embedding pipeline. Most commonly used downstream tasks to evaluate embeddings, their evaluation metrics and popular datasets are highlighted. Finally, we present the open-source libraries for network embedding.

## 1. Introduction

A massive amount of data is recently represented in a wide range of online applications taking the form of complex systems. Networks are powerful data structures used to explore and model complex systems in real-world including online social networks, like Facebook and Twitter, academic networks, like DBLP and Google Scholar, as well as biological networks, like Protein-Protein interaction networks. Analyzing these networks can provide insights on how to take profit from the hidden information in networks. For that reason, many machine learning algorithms look for making predictions and discovering new patterns using network data as vector features. For instance, one may predict the role of a protein in a biological network or recommend new friends to a user in a social network. While traditional machine learning algorithms use a vector of features as input, network data are composed of relations that encode additional valuable information allowing inference among nodes. Furthermore, network data can be very hard to cope with due to their very large-scale (networks with billion of nodes and edges), heterogeneous structure (networks with different types of nodes and edges) as well as auxiliary information related to nodes and/or edges.

Great challenges exist in analyzing these massive networks with traditional machine learning algorithms, which exploit feature vectors as the input and cannot handle graph data directly. Hence, the first challenge to deal with is to find an effective representation of network data that represent networks concisely so that machine learning based tasks, like link prediction and node classification, can be applied efficiently in both time and space (Cui, Wang, Pei, & Zhu, 2018). For instance, in the case of link prediction task, one might want to encode the number of common friends between two nodes or capture the structure of the node's global or local neighborhood. Another challenge from machine learning perspective is that there is no straightforward way to encode this high-dimensional data about the network structure into a feature vector.

To deal with the above presented challenges, a surge of approaches seeking to automatically *learn* representation for network data have been proposed (Figueiredo, Ribeiro, & Saverese, 2017; Grohe, 2020; Grover & Leskovec, 2016; Ou, Cui, Pei, Zhang, & Zhu, 2016; Perozzi, Al-Rfou, & Skiena, 2014). The idea behind *Network Representation Learning (NRL),* also known as *network embedding,* approaches is to embed nodes, edges or the entire (sub)graph into a low-dimensional vector space $\mathbb{R}^d$. This mapping should preserve network properties so that, after an optimization step of the learned embeddings, geometric relationships in the embedding space reflect the structure of the original network. Learned embeddings are then used as input to machine learning algorithms to extract useful information when building classifiers or other predictors.

Several previous surveys (Ali, Kefalas, Muhammad, Ali, & Imran, 2020; Goyal & Ferrara, 2018; Kazemi et al., 2020; Khosla, Setty, & Anand, 2019; B. Li & Pi, 2020; Moyano, 2017; Q. Wang, Mao, Wang, & Guo, 2017; Carl Yang, Xiao, Zhang, Sun, & Han, 2020; D. Zhang, Yin, Zhu, & Zhang, 2018; Z. Zhang, Cui, & Zhu, 2020) existing in the recent literature are related to our paper. A comparative analysis of these surveys and our proposed survey are presented in table 1.

**Table 1: Comparison between different NRL related surveys**

| Survey Ref. | Coverage | Network types | NRL Models | Detailed Process of NRL | Key findings | Limitations |
|---|---|---|---|---|---|---|
| (Goyal & Ferrara, 2018) | 2000-2017 | Homogeneous networks | Various | Not provided | • Proposes a taxonomy of graph embedding approaches<br>• Provides various applications of embedding and their respective evaluation metrics | • Covers traditional models and does not treat new models of NRL<br>• Does not provide different types of network data<br>• Does not provide the detailed process for NRL |
| (Kazemi et al., 2020) | 2007-2020 | Heterogeneous/ dynamic networks | Encoder-decoder perspective | Not provided | • Focuses on recent representation learning techniques for dynamic graphs<br>• Categorizes existing models using an encoder-decoder framework | • Focuses only on NRL models for dynamic graphs<br>• Does not provide a detailed process for NRL |
| (Khosla et al., 2019) | 2009-2019 | Non-attributed networks | Various | Not provided | • Proposes a framework that casts a variety of unsupervised approaches into a unified context-based optimization function<br>• Systematically classifies studied models based on their similarities and differences | • Considers only unsupervised approaches<br>• Does not consider different types of network data<br>• Does not cover a detailed process for NRL |
| (B. Li & Pi, 2020) | 2010-2020 | Homogeneous and heterogeneous network | Various | Not provided | • Classifies network embedding in the framework of homogeneous and heterogeneous network<br>• Presents various applications related to network embedding and their respective evaluation | • Does not consider other types of networks such as attributed graphs<br>• Does not cover a detailed process for NRL |
| | | | | | metrics | |
| (Z. Zhang et al., 2020) | 2009-2018 | Homogenous networks | Various | Not provided | • Categorizes NRL models based on two settings: unsupervised setting and semi-supervised settings<br>• Categories existing models based on their architectures and training strategies | • Does not tackle different types of network data<br>• Does not cover a detailed process for NRL |
| (Ali, Kefalas, et al., 2020) | 2010-2020 | Homogeneous and heterogeneous networks | Deep learning based | Not Provided | • Categorizes deep learning based citation recommendation models using six criteria<br>• Presents a comparative analysis of citation recommendation models | • Does not cover a detailed process for NRL<br>• Does not cover other downstream tasks other than citation recommendation |
| **This survey** | 2010-2021 | Homogeneous/ heterogeneous, non-attributed, attributed, weighted/un-weighted, directed/undirected networks | Various | Provided | • Provides a comprehensive view of the reasons behind the emergence of network embedding<br>• Covers different components of a NRL pipeline including input setting, context graph generation, NRL used models, and different types of output settings<br>• Presents possible downstream tasks, for learned embeddings, along with their evaluation metrics, used datasets, and open-source Python libraries for existing NRL models | |

The first survey on network representation learning is introduced by Moyano (2017) in 2017. This work reviews few representative approaches for NRL and highlights some basic concepts, such as hyperbolic embedding, stochastic embedding, and neural network embedding, related to NRL and its relations to dimensionality reduction techniques, deep learning and network science. Existing related works are reviewing network representation learning models on different types of graphs including heterogeneous networks (Carl Yang et al., 2020), dynamic networks (Kazemi et al., 2020), and knowledge graphs (Q. Wang et al., 2017). Other surveys are reviewing NRL algorithms considering one single taxonomy like NRL based on deep learning models (Ali, Kefalas, et al., 2020; W. L. Hamilton, Ying, & Leskovec, 2017; Z. Zhang et al., 2020) and unsupervised NRL (Khosla et al., 2019). Moreover, Goyal and Ferrara (2018) and Zhang et al. (D. Zhang et al., 2018) both of them classify different proposed network embedding algorithms from a methodology perspective based on the type of model used to learn representations. Still, most of reviewed NRL algorithms cited in these surveys preserve network structure. Lately, Cui et al. (2018) reviewed NRL algorithms exploiting other side information, such as node attributes, to learn efficient embeddings. A recent survey conducted by Li and Pi (2020) reviews existing network embedding algorithms and classify them based on both network input type and embedding output type. All previously cited surveys have the following defects more or less. Firstly, some of these reviews focus on one single taxonomy to classify existing works and they treat the problem of network embedding in different perspectives separately. Secondly, most of studies do not make difference between various types of network data input as well as network embedding output. Majority of works focus on one single type of network data input, i.e. homogeneous network, and one single type of network embedding output, i.e. node embedding. Thirdly, there is a lack of summary on providing available datasets for network embedding on real-world information networks, popular used downstream tasks, evaluation metrics used in each downstream task to evaluate learned embeddings along with open-source network embedding frameworks.

In this respect, the present work is intended to provide a comprehensive and systematic review of network representation learning by highlighting the reasons behind its emergence and providing a detailed NRL process through the classification of different methods used in each step of network embedding pipeline. Accordingly, in order to provide a comprehensive view of network representation learning for new researchers and foster the future researches in the field, we have formulated two main questions to which we try to provide plausible answers through the undertaken survey, namely:

- What are the reasons behind network embedding emergence?
- What types of settings as well as models are used in each step of network embedding pipeline?

In the following section, we describe the methodology used to conduct the present survey. In section 3, we provide an overview of NRL artificial intelligence related context and more precisely representation learning in overcoming the problem of feature engineering as a labor-intensive task to create a new data representation for machine learning based tasks. Representation learning is not a novel approach and embedding data into a low-dimensional space dates back to 1980 (Rosenberg, 1980). Therefore, we briefly review some of the traditional graph embedding approaches as a part of dimensionality reduction techniques. In section 4, as natural language processing (NLP) domain was among the earliest domain of application of representation learning, a brief history is presented as well as the famous Word2vec model, a neural network language model, for word representation learning. Showing excellent potential in NLP through Word2vec model, word representation learning has been widely used outside of NLP to analogically represent other types of information and hence many novel vector representations under the umbrella of "*Anything2vec*" have been emerged and discussed in section 5. Then, in section 6, we first focus on challenges of traditional graph embedding when applied to complex systems and we introduce the formal definition of information network as well as its properties and provide a formal problem definition of network embedding. Next, network embedding pipeline along with categories of models used in each step are presented in section 7. In fact, four main parts are of importance in a network embedding framework: type of the input graph (homogeneous, heterogeneous or attributed), used approach to define context graph (random walk based approaches, adjacency based approaches and others, e.g. similarity based approaches), used network representation learning approach (skip-gram based approaches, matrix factorization based approaches, deep learning based approaches, and others), and finally the type of the network embedding output (node embedding, edge embedding, substructure or hybrid embedding and whole graph embedding). In section 8, we review some network embedding evaluation tasks, like link prediction, node classification and network visualization. In addition, available network embedding libraries are presented in section 9 and section 10 concludes the survey with a discussion of some possible future research directions.

## 2. Survey applied methodology

In this work, we present a systematic review for network representation learning. An academic search is conducted through various important academic libraries based on a set of relevant predefined terms that can figure out either in the title or in the abstract of a given article. Used methodology is further detailed in the following subsections.

### 2.1. Search strategy: academic libraries and search terms

In recent years, network representation learning has witnessed a surge of research in the well-known artificial intelligence and data mining journals and conferences including

TKDE, AAAI, TPAMI, and IJCAI. Our search strategy takes into consideration the entirety of published journal articles and conferences from four main academic libraries dealing with the computer science area, namely: Elsevier Science Direct, Springer Link, IEEE (Institute of Electrical and Electronics Engineers) Xplorer, and ACM (Association for Computing Machinery). The search focuses on retrieving related works about network representation learning and network embedding. To this end, a set of predefined terms are used, more particularly, "network representation learning", "network embedding", "graph embedding" as equivalent terms in the same search area. Network representation learning covers many sub-directions such as node embedding, edge embedding, sub-graph embedding, heterogeneous information embedding, attributed network embedding, and so forth. Hence, we try sometimes to leverage additional key terms related to these sub-directions of NRL to get more relevant articles.

## 2.2. Study inclusion and exclusion criteria

The criteria, used to identify convenient works satisfying the research question requirement among all studies collected from the list of mentioned academic libraries, are as follows:

- The study should be published between the two following years: 2010 and June 2021
- The study must be written in English.
- Retrieved article must tackle one of the sub-directions of network embedding or it may discuss important axis of network embedding evolution in terms of taxonomies, problem settings, and evaluation downstream tasks along with the popular used datasets, evaluation metrics, and open-source libraries.

Table 2 summarizes the number of retrieved articles, meeting the inclusion criteria, from each electronic database.

**Table 2:** Number of retreived articles per academic library

| Academic Library | Retrieved articles meeting inclusion criteria |
|---|---|
| Elsevier | 112 |
| ACM digital library | 98 |
| Springer Link | 130 |
| IEEE Xplorer | 170 |

Articles meeting inclusion criteria are collected based on a set of predefined keywords, mainly network representation learning and network embedding. All the retrieved articles from the four electronic databases include at least one keyword from predefined

keywords either in the title of the document or in the abstract section. Figure 1 shows the number of published papers on this topic from 2010 to June 2021.

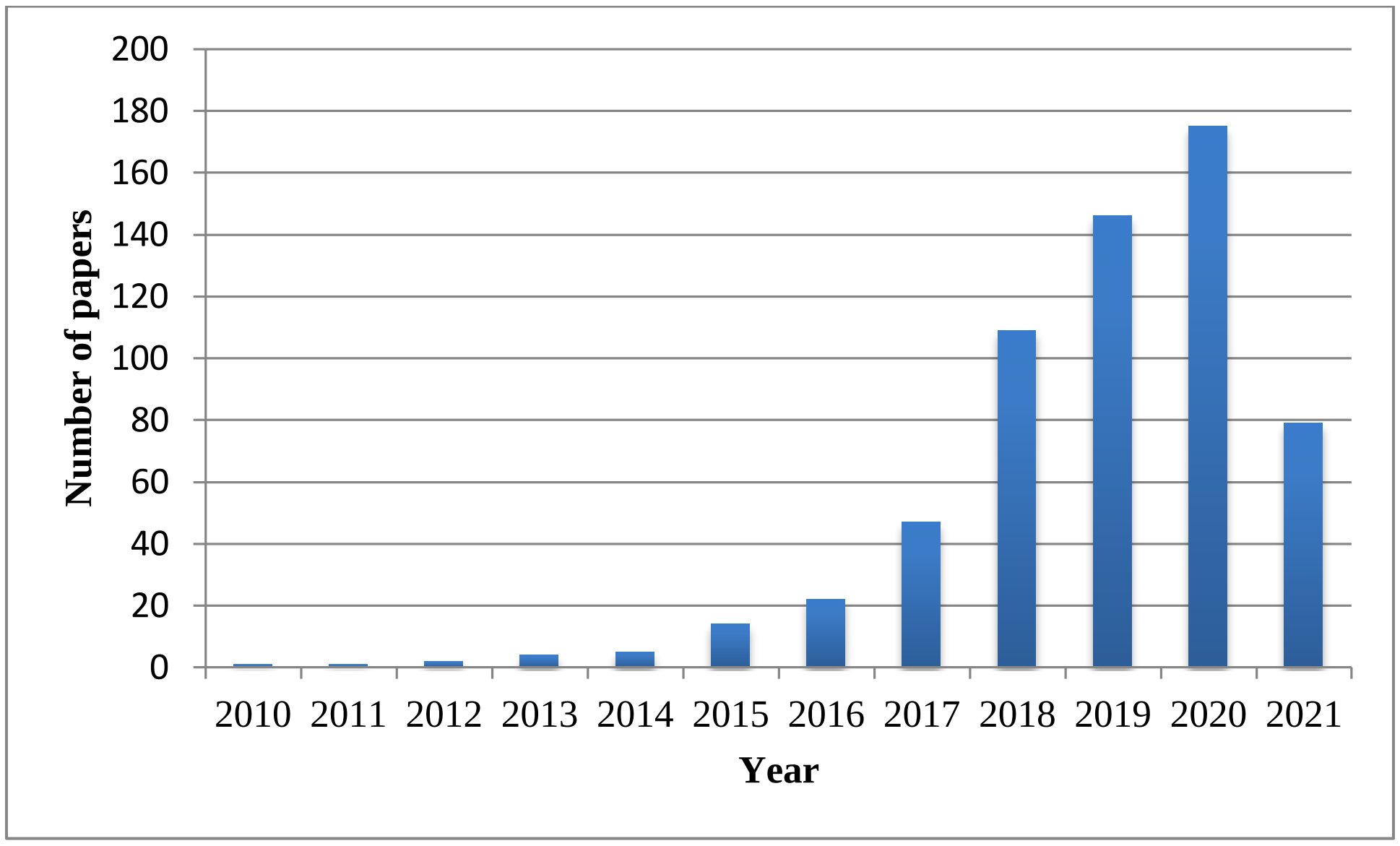

**Figure 1: Statistics of papers focusing on Network Embedding from 2010 to June 2021**

## 3. Representation learning: artificial intelligence related context

Since the invention of the computer age by Alan Turing in 1950, the ultimate goal of the Artificial Intelligence (AI), that a machine can have a human-like general intelligence and interpret world as human do, is one of the most ambitious ever proposed by science. It focuses on intelligent agents that have human intellectual characteristics, behaviors, learning from past experiences and effectively solve problems.

Due to its capabilities to develop intelligent systems, AI has been used into several major fields including transportation, finance, national security, agriculture, and education. The application of AI is currently ranging from the commercial and service industries to the manufacturing and agricultural fields, which makes the AI general technology more prominent and spread. For instance, in the financial field, AI is currently utilized in fraud detection (West & Bhattacharya, 2016), voice and pattern recognition (Eljawad et al., 2019), and investment strategies to efficiently adapt regulatory and infrastructural changes as well as reducing errors. In educational field, the combination of AI and education give a chance for the teaching model to be more intelligent, the educational resources become more dynamic and open, and the learning mode move to be more adaptive to the capabilities of learners (Mosenthal, 1981). Another example of AI implication in real life is in the healthcare domain which witnessed significant investments by many governments and many technology companies (Reddy, Fox, &

Purohit, 2019). For instance, AI-enabled medical devices are actively introduced in the market by the United States Food and Drug Administration[1].

In computer science domain, there are plenty of examples where the use of artificial intelligence is blooming and expanding in all its format such as knowledge engineering and Natural Language Processing techniques ranging from statistical machine learning algorithms, predictive models to various algorithmic approaches. Transfer learning or domain adaptation, object recognition, speech recognition and signal processing (Bengio, Courville, & Vincent, 2013) are other examples of AI and machine learning applications. The performance and success of machine learning algorithms are heavily dependent on the choice of data representation on which they are applied. For that reason, in the deployment of machine learning algorithms, much of the effort goes into the design of feature extraction, preprocessing pipelines, and data transformations that result in a representation of the data that can support effective machine learning.

### 3.1. Feature engineering

Feature engineering is a way about creating a new data representation (or features) from raw data using data mining techniques. It covers the selection of a subset of informative features (dimensions) that one may obtain a representation enabling a specific task. This hand-crafting feature engineering usually requires a deep understanding on domain knowledge. For example, in the case of disease outbreak, experts manually define dictionaries of terms related to the disease, e.g., symptoms and medications, to identify notes that assert the presence of it (Uzuner, 2009). Hand-crafting feature engineering methodologies count heavily on human design and implementation and they are almost of time based on an educated guess of what will be useful (Prusa & Khoshgoftaar, 2016). For that reason, feature engineering is labor-intensive, especially when the raw data are high-dimensional and non-linear, and hence cause the weakness of machine learning algorithms. As a result, machine learning algorithms are unable to extract all the juice from raw data and hand-crafted features are usually designed for specific task and do not generalize for over different machine learning algorithms (Grover & Leskovec, 2016). It would be highly preferred to make machine learning algorithms separate from feature engineering so that innovative applications could be built faster and to progress more towards artificial intelligence.

### 3.2. Representation learning

Due to the complexity, high-dimensionality and non-linearity of data especially in the era of "big data", there is a need to find a convenient *low-dimensional* representation of data, which means that the learned dimension is significantly smaller than the dimensionality of the observation space, that reflects the main explanatory factors for the observed variations in the raw data and may reveal relations in the data that are harder to manually identify. The possibility to automatically *learn* data representation is at the center of the efforts that push the research forward in this area.

[1] https://www.scipol.org/content/fda-moves-encourage-ai-medicine-drug-development

The manifold hypothesis (Bengio et al., 2013) is an essential hypothesis for representation learning and according to which real-world data presented in high dimensional spaces are expected to concentrate in the vicinity of a manifold $M$ of much lower dimensionality $d_M$. A natural coordinate system is provided by this manifold to the representation being learned. Proposed approaches for representation learning cover a wide range of applications. Representations learning on image (Guo et al., 2016; Krizhevsky, Sutskever, & Hinton, 2012; Wright et al., 2010) , speech (Dahl, Yu, Deng, & Acero, 2011; Graves, Mohamed, & Hinton, 2013), natural language processing (T. Chen & Sun, 2017; Collobert & Weston, 2008), and network data (Goyal & Ferrara, 2018; F. Zhou et al., 2018) are among others. Commonly, a good representation needs to have some key components to correctly prescribe the intricacies of real-world data (D. Wang, Cui, & Zhu, 2016a). It should also preserve data structures in the way that similar data points should be integrated closer in the representation space. As they can be efficiently processed in both time and space, learned representations can benefit a variety of applications such as link prediction, node classification, recommendation systems, and even more biological related applications. In the link prediction task, if one might want to predict, for instance, missing relationship or biological interactions between two nodes in a given network, it needs to extract a set of features capturing the nodes' neighborhood. Such method may fail to encode inherent or hidden information in the network and hence leads to low accuracy in the prediction task. Consequently, learned representations are able to capture hidden information in a given data and thus resulting in improved model accuracy.

### 3.3. Dimensionality reduction, linear and non-linear models

Real-world data, such as speech signals, digital images, and texts, are usually characterized by high-dimensionality which needs to be reduced. Ideally, the reduced dimensionality has to encode the minimum number of parameters needed to account for the observed properties of the data.

Dimensionality reduction has a long history. Its proposed models fall broadly into two categories: linear and non-linear models. There are several linear learning techniques used for dimensionality reduction to describe data variability and similarity such as Principal Component Analysis (PCA), Linear Discriminant Analysis[2] (LDA), Locality Preserving Projections [3](LPP), Multidimensional Scaling[4] (MDS), and their extensions.

[2] Linear Discriminant Analysis is a supervised dimensionality reduction technique that takes labels into consideration to map data into a low-dimensional space.

[3]Locality Preserving Projections is a linear approach for dimensionality reduction trying to capture the non-linear manifold structure of the data and finding an embedding that preserves local information.

PCA is the most commonly used linear dimensionality reduction method and it computes a hyperplane that passes through data points as best as possible in least square-sense. For the second category of dimensionality reduction, a large number of techniques have been proposed to deal with complex non-linear data as it is the case of real-world data that is likely to form a highly non-linear manifold. Isometric Feature Mapping [5](ISOMAP) (Tenenbaum, De Silva, & Langford, 2000), Local Linear Embedding[6] (LLE) (Roweis & Saul, 2000), Local Spline Embedding[7] (LSE) (Xiang, Nie, Zhang, & Zhang, 2008), stochastic neighbor embedding[8] (SNE) (G. E. Hinton & Roweis, 2003) as well as manifold learning techniques (Lee & Verleysen, 2007) represent examples of non-linear models. All the above-mentioned models count as traditional and classical models token as input a set of identical and independently distributed data points to build an affinity graph, and then embed the affinity graph into a new space having much lower dimensionality (D. Zhang et al., 2018). Dating back to the early 1980s (Rosenberg, 1980), the notion of *graph embedding* is emerged, and it attempts to place the data in a low-dimensional space so as to optimally preserve original neighborhood identity.

## 4. Word representation learning ancestor

NLP has token profit from the field of artificial intelligence to go behind the understanding of word senses. To understand the reasons behind the emergence of word embedding, a brief history is presented in the following section.

### 4.1. Background

Words have traditionally been encoded as discrete symbols which cannot be useful to identify semantic and syntactic similarities between words. Conceptually, such approach is equivalent to "one-hot encoding" (Socher, 2014) representation which uses a high-dimensional zero-one vector with dimensionality equal to the vocabulary size. In fact, for each word vector $v_w$, all entry values will be set to "**0**" except one entry value corresponding to the position of the word *w* in the vocabulary will be set to "**1**". With such an encoding, in one hand, all words have independently been encoded from each other and hence there is no easy way to identify similar words in vector space. In the other hand, assuming that we have a vocabulary with 100,000 terms and thus each word vector has a size of 100,000 dimensions. Clearly, one-hot encoding representation uses many zero-values which do not contain any information leading to the waste of sparse space.

[4] Multidimensional Scaling is a linear dimensionality reduction technique that creates a map displaying the relative positions of a number of objects, given only a table of the distances between them.

[5] Isometric Feature Mapping finds the map that preserves the global, nonlinear geometry of the data by preserving the geodesic manifold interpoint distances.

[6] Local Linear Embedding attempts to discover nonlinear structure in high dimensional data by exploiting the local symmetries of linear reconstructions.

[7] Local Spline Embedding is a nonlinear dimensionality reduction technique that represents each data point in different local coordinate systems and aligns them in a global coordinate through compatible mappings.

[8] Stochastic neighbor embedding is another non-linear technique used to place the objects in a low-dimensional space so as to optimally preserve neighborhood identity, and can be naturally extended to allow multiple different low-d images of each object

To overcome both syntactic and semantic similarity between words as well as the sparse wasted problem of space, word representation learning takes place for acquiring such representations based on the distributional hypothesis of Harris (1954) affirming that words in similar contexts have similar meanings. Based on this distributional hypothesis, many word representation approaches are proposed to encode words in continuous, low-dimensional and dense vectors known as distributed term representations or word embeddings. The concept of word embedding has a long history (Bengio, Ducharme, Vincent, & Jauvin, 2003; DE, Hinton, & Williams, 1986; Elman, 1990; G. E. Hinton & others, 1986; G. Hinton, McClelland, & Rumelhart, 1986; Z. Liu, Lin, & Sun, 2020). It was first introduced on 1986 by Hinton et al. (1986) to learn distributed representation of concepts and it was advocated since the emergence of neural networks (DE et al., 1986; Elman, 1990). In work done by Bengio et al. (2003), to represent words with a distributed continuous vectors, authors estimated the joint probability of sequences of words through the estimation of the conditional probability of the target word given a few previous words representing the context. A neural network with a particular structure is used to define this conditional probability function. This function takes as input a word symbol and first embedded it into a Euclidean space by learning to associate a real-valued “feature vector” to each word. Furthermore, the use of neural networks for language modeling is not new (Miikkulainen & Dyer, 1991; Xu & Rudnicky, 2000) and it is inspired from proposals of character-based text compression using neural networks to predict the probability of the next character (Schmidhuber & Heil, 1996).

The idea of word embedding has been applied to statistical language modeling with considerable success (Bengio et al., 2003). Several neural network language models were proposed (Bengio et al., 2003; Krbec, 2006; Mikolov, Kopecky, Burget, Glembek, & others, 2009), where the word vectors are learned using neural networks with single hidden layer and then fed into neural network language models to train them. Other types of language models are the well-known Latent Semantic Analysis (LSA) and Latent Dirichlet Allocation (LDA).

To the best of our knowledge, the work performed by Morin and Bengio (2005) is the first research work that used at the first time the term of ‘*word embedding*’ in this context. They introduced a new architecture for speeding-up neural networks with a huge number of output classes using a Hierarchical Softmax[9] layer. A recent research work presented by Mikolov et al. (2013), which racked up plenty of citations till the time of writing, completely changed the entire landscape of NLP. Indeed, the Word2Vec[10] model was released carrying out simple but efficient approaches for inferring dense vector representation of words using contextual information from large unlabeled corpora with basically a very large and even unlimited size of word vocabulary.

[9] Hierarchical Softmax is a computationally efficient approximation of the full softmax. To obtain the probability distribution, it evaluates only about $log_2(N)$ nodes instead of evaluating N output nodes in the neural network.

4.2. Word2Vec model: inspiration source for network embedding

Word2Vec model (Mikolov et al., 2013) is a neural network language model that consists of three-layers: one input layer, one output layer and one hidden layer with no activation function. It is based on efficient architecture to provide interesting semantic properties of words from the *context* in which they are used. Large datasets are used when training the Word2Vec model in order to represent the semantic and syntax of words accurately with the purpose that similarity between words can be measured effectively. Two different but related variants were proposed: skip-gram and continuous bag-of-words (CBOW) models. Given the current word, skip-gram model predicts the surrounding context based on contextual window, while CBOW predicts the current word given its context.

Figure 2 shows both architectures of skip-gram and CBOW models of Word2Vec. For skip-gram model, the dimensionality of the input layer is equal to the size of vocabulary $V$ and contains the one-hot vector $\vec{w}_t$ of the current word. The output layer has the same size as the input layer and it encodes the one-hot vectors $\vec{w}_{t-i}$ and $\vec{w}_{t+i}$, where $i \in [1..c]$ and $c$ represents the context window, of words in the surrounding context. The probability that word $w_j$, where $j \in [t-c..t+c]$ is in the context of word $w_t$ is defined as the softmax of their vector product:

$$p(w_j|w_i) = \frac{\exp(v_{w_i} v_{w_j}^T)}{\sum_k^{|V|} \exp(v_{w_i} v_{w_k}^T)} \qquad (1)$$

where $v_{w_i}$ and $v_{w_j}$ represent the input and output vector representations of words $w_i$ and $w_j$.

The skip-gram model aims to maximize the probability of the neighbors of word $w_t$ in a given sequence of words conditioned on its embedding $\emptyset(w_t)$ as follows:

$$\max_{\emptyset} \mathrm{P}(\{w_{t-c}, .., w_{t+c}\} \backslash w_t | \emptyset(w_t)) = \prod_{j=t-c, j\neq i}^{t+c} \mathrm{P}(w_j | \emptyset(w_t)) \qquad (2)$$

The CBOW model is similar to the skip-gram model but its ultimate goal is to predict the target word given the surrounding context.

The skip-gram model is generalized for network representation learning and prior works on network embedding are based upon this neural network language model. By making an analogy between natural language sentences and random walk sequences, various studies (Figueiredo et al., 2017; Grover & Leskovec, 2016; Perozzi et al., 2014) have inspired the idea of skip-gram model to learn latent representations of nodes in a given network.

[10] https://code.google.com/archive/p/word2vec/

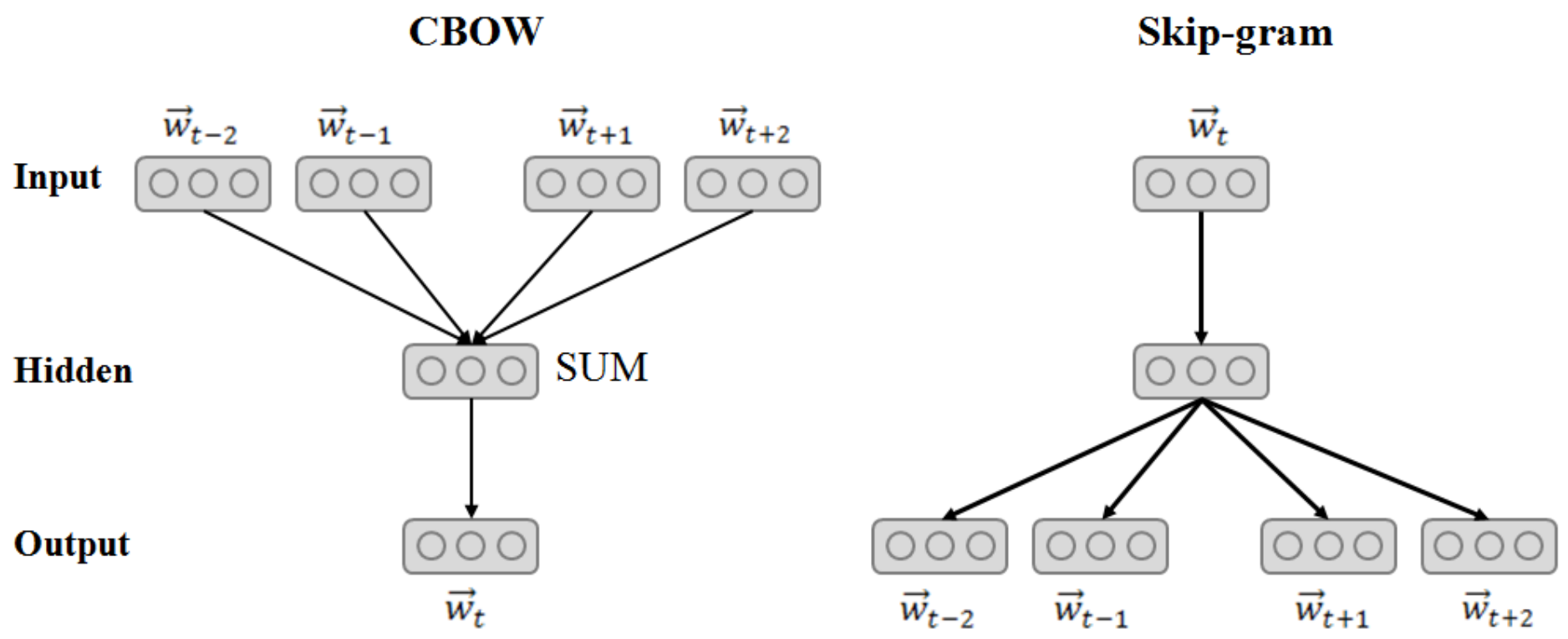

**Figure 2:** General architectures of CBOW and Skip-gram models of word2Vec (Mikolov et al., 2013)

There are other models used to learn word representation: FastText and GloVe. While Glove has been investigated for the first time by Brochier et al. (2019) to learn node embedding for different types of information networks, FastText has not been investigated yet for network embedding. To learn word representations, Glove stands for "Global Vectors" and it captures global and local statistics of a corpus of documents using its co-occurrence matrix. Inspired by Glove, Brochier et al. (2019) proposed a matrix factorization approach for network embedding where nodes of the network represent documents of the corpus.

## 5. Anything2vec trends

One of the key challenges of typical machine learning algorithms is dealing with the high-dimensionality of data, given that we are actually living in the era of big data. This high-dimensionality problem has posed itself in many disciplines like NLP, image recognition, and social network analytics. Accordingly, showing excellent potential in NLP through word2vec model, word representation learning has been widely used outside of NLP to represent other types of information, e.g., papers or authors in network citations (Jian Tang et al., 2015). The use of neural network based architecture, with a large number of parameters, to automatically learn a high quality continuous representation of interesting concept such as a word, an event, a document, node in a graph, a user, a hashtag in social networks, news, and structure of a graph to name few, is the key novelty of representation learning of concepts into low-dimensional space. These learned low-dimensional vectors play an important role in extracting proper representations from complex data for various analyzing tasks. Consequently, a significant amount of progresses have been made towards this theory of vector embeddings in structured data.

A *vector embedding* (Grohe, 2020) for a class $X$ of objects is defined as a mapping function $f$ from $X$ into the so-called latent space $R^d$, a real vector space of finite dimension $d$. The ultimate goal is to define a vector embedding so that semantic relationships between the objects in $X$ are reflected by geometric relationships in the latent space. The most important feature with vector embedding is that similar objects in $X$ have to be mapped to vectors close to one another on the latent space. Similarities between objects of $X$ and semantic relationships cannot be quantified precisely and may be application dependent. Vector representations can be learned to carry out good results for solving a specific machine learning tasks known as *downstream tasks*. There is a wealth of studies that focus on learning vector representations for specific tasks. Le & Mikolov (2014) proposed doc2vec or Paragraph Vector as a simple extension of word2vec to learn document-level embeddings which have been fed into machine learning algorithms for classifying documents in document retrieval, web search and spam filtering applications. Ganguly et al. (2016) presented Author2Vec model which learns low-dimensional representations for authors by combining information from scholars' co-authorship network and the content of abstracts. Other recent works have introduced medical representation learning models to capture the rich latent relationships between medical concepts, e.g., Med2Vec (Choi et al., 2016) and Char2Vec (Hussain, Moosavinasab, Sezgin, Huang, & Lin, 2018) models. The former is based on neural network architecture to learn the interpretable representations for both medical codes and visits from Electronic Health Records datasets. Char2Vec (Hussain et al., 2018) introduced a character-level semantic indexing model for learning the semantic embedding of rare and unseen words in the biomedical literature. Last cited related works are just few models among a rapidly growing number of *anything2Vec modeling* algorithms, cited in *appendix 1,* for representing objects as vector embeddings.

In the following two sections, we present formal definitions for some basic concepts of network representation learning and point out network embedding problem settings as well as different types of models used in each step of network embedding pipeline.

## 6. Network Embedding

Traditional graph embedding models highlighted above, originally studied as dimensionality reduction techniques, are generally constructed from feature represented datasets, like image datasets, having a clear grid structure where proximity between nodes encoded by weights of the edges is well defined in the original space. In fact, naturally formed networks exist in a wide variety of real-world scenarios, e.g., online social networks, citation networks, knowledge networks, e-commerce networks, biological networks, and so forth. Different types of informational objects interconnected between them through various types of links or edges, are interacting with each other, forming large, sophisticated and interconnected networks. Taking the example of online social networks, nodes in these networks represent users, images, check-ins, videos, groups, etc, and complex relationships could be formed between them, e.g., social relations. Without loss of generality, these networks are named as *information networks*. Hence, analyzing such networks adds insights into how to effectively take profit from hidden information in them. In these information networks, proximities among nodes are not explicitly and clearly defined and the application of traditional graph embedding approaches to large real-world networks is challenging due to many reasons (Cui et al., 2018; Z. Zhang et al., 2020):

- *High computational complexity and large-scale networks*: Real networks like social networks or citation networks are composed of millions or billions of nodes and edges. Therefore, designing scalable models that are characterized by a linear time complexity with respect to the graph size is not a trivial problem. In fact, shortest or average path length is generally the most used metric to represent distance between two nodes. Computing such distance using traditional graph embedding approaches needs the enumeration of possible paths between two nodes and result in high computational complexity.
- *Diversity of real-world networks*: real-world networks are complex systems composed of diverse types of nodes as well as edges, i.e., properties. They can be homogenous or heterogeneous, directed or undirected, weighted or un-weighted, attributed or not attributed graphs. Furthermore, many downstream tasks can be proposed to perform network processing and analysis in order to infer insights from graph data. These downstream tasks are widely varying from node-based problems, e.g., node classification and link prediction, to graph-based problems such as graph classification. As a result, new deep learning models are required to deal with specific tasks since already existing approaches of graph embedding cannot be applied to such problems due to their inefficiency to handle such challenges.
- *Graph properties preserving, attribute and structure*: Information networks are characterized by a set of naturally formed nodes and edges so that proximities among nodes are not well defined. Therefore, network structure and other side information such as node attributes and/or node labels should be preserved when learning the embedding representations. For that reason, a context representation has to be defined under graphs to measure the similarity information among nodes so that similar nodes in the original network are embedded closer in the embedding space. In this respect, as part of dimensionality reduction techniques, traditional graph embedding approaches, such as ISOMAP, LLE, LSE, and so on, are conceived for graphs with clear grid structure. These graphs are characterized either by linear 1D structure, like text and speech graph, or by a fixed 2D structure, like image dataset, in where graph size, node ordering, and proximity between nodes are well defined. However, real-world information networks do not match such properties as they present more complex topographical along with an arbitrary size and node ordering. As a consequence, traditional graph embedding techniques are not applicable for current large information networks and hence there is a need for new techniques to quantify similarity in such real-world graphs.

Due to these challenges, tremendous efforts have been made in this area resulting in the development of novel representation learning approaches for real-world networks under the umbrella of network representation learning or network embedding. Network embedding aims to learn dense or latent vectors for network nodes where relationships between nodes are captured by the distance between them in the vector space. Topological and structural features of nodes are encoded into their low-dimensional representations. The learned embedding space can effectively catch the goal of network inference to discover hidden semantics among different components of the network.

***Definition 1***: Information Network

Given a graph $G = (V, E)$, where $V = \{v_1, v_2, .., v_{|V|}\}$ represents the set of vertices and $E = \{e_{ij}\} \subseteq V\ x\ V$ defines the set of edges among nodes, with an objective mapping function $\varphi: V \rightarrow T$ and an edge type mapping function $\emptyset: E \rightarrow L$. Each node $v$ pertains to one particular object type $\varphi(v) \in T$ and each edge $e$ pertains to one particular edge type $\emptyset(e) \in L$.

- **Property 1**: If the types of objects $|T| = 1$ and types of the edges $|L| = 1$, the information network is called *homogeneous information network*; otherwise it is *heterogeneous information network*.
- **Property 2**: An adjacency matrix $A \epsilon R^{|V|} \times R^{|V|}$ can be associated with $G$ where $A[i][j] \neq 0$, if $(v_i, v_j) \in E$, otherwise $A[i][j] = 0$. In that case, the graph G is denoted as *weighted graph*. When $A[i][j]$ is equal to 1 for $i$ and $j$, the graph $G$ is denoted as *un-weighted* graph.
- **Property 3**: G is *undirected* graph if $A[i][j]$ is symmetric and $A[i][j] = A[j][i]$ for all $i$ and $j$. If the order of the nodes in edges is important, the graph G is *directed*. In such case, for a link $(v_i, v_j)$, $v_i$ represents the source of the edge and $v_j$ is the target of the edge.
- **Property 4**: The graph G is an *attributed graph* if for each node *v* is associated a set of properties representing its characteristics. If all nodes have the same set of attributes, all attributes values are represented through a matrix $X$ whose $i^{th}$ row corresponds to the attributes values of node $v_i$.

***Definition 2:*** Network embedding

Given an information network $G = (V, E)$, the task of network embedding is to learn a mapping function $f: v_i \rightarrow y_i \in R^d$ where $i \in \{1, .., |V|\}$, with $d \ll |V|$ and $d$ defines the number of dimensions of the real-valued vector as well as the dimension of the latent space in which graph property is preserved as much as possible, and $y_i$ is the learned vector of the node $v_i$ supposed to contain the necessary information about the node and its relations to enable making predictions about it. The embedding vectors are expected to preserve the various features of the information network as rich as possible.

- **Property 1:** If $G$ is homogeneous information network, the objective function $f$ should preserve the various features of the network as much as possible; otherwise the network G is heterogeneous and the objective function $f$ embeds the different types of network nodes into low-dimensional space and guarantees that nodes sharing similar semantics in $G$ have similar latent representations $f(v)$.
- **Property 2:** If $G$ is weighted, the objective function $f$ should guarantee that nodes having higher weights on edges between each other have to be embedded closer in the embedding space than others with which they have a low weights.
- **Property3:** Similarly, if the network $G$ is directed, the mapping function should guarantee that nodes which are connected with other nodes in both directions must be embedded closer in the latent space than other nodes with which they are connected in one direction.
- **Property 4:** If $G$ is an attributed network, the mapping function $f$ embeds each node into a low-dimensional representation based on the adjacency matrix $A$ and the attribute matrix $X$, such that $f$ is expected to preserve the proximity existing between the topological structure and nodes attributes.

## 7. Network embedding pipeline and problem settings

In this section, an overview of the general architecture of a network embedding system is discussed. It mainly consists of 4 parts: network embedding input, context definition and generation, latent network representation learning models, and network embedding output. In network embedding input part, we start by distinguishing between different types of networks and then shed lights on how these properties can help building better embeddings and capture semantics among nodes in given graphs. Following this part, the notion of "context graph" is defined and different approaches used to generate it are introduced. After that, different models used to learn latent representations are discussed and categorized into 3 parts; skip-gram based approaches, deep learning based approaches, and miscellaneous, e.g., LINE. A network embedding system can produce several types of outputs ranging from node embedding to edge and whole-graph embedding. These learned representations can be introduced into various downstream tasks for prediction, recommendation or clustering. Figure 9 presents the categorization of different settings and approaches used in each step of network embedding pipeline. Finally, table 3 presents the classification of different NRL models used in this survey in each step of network embedding pipeline.

7.1. Network embedding input

The input of network embedding system is a graph which can be categorized mainly into three types: homogeneous, heterogeneous and graph with auxiliary information.
*Homogeneous graph:* both nodes and edges have one single type respectively and edges between nodes can be weighted (or directed) or un-weighted (or undirected). The most basic and commonly used network embedding input is the undirected and un-weighted homogeneous graph in which only structural information is available (Perozzi et al., 2014; Pimentel, Veloso, & Ziviani, 2017; X. Wang et al., 2017). Exploiting the weights and directions of edges can intuitively provide more additional information about the graph helping to represent the latter more accurately in the latent space. For instance, let's take the example of a graph composed of three nodes $\{ v_1, v_2, v_3 \}$ with two edges $e_{v_1,v_2}$ and $e_{v_2,v_3}$ having the weights $w_{v_1,v_2} = 1$ and $w_{v_2,v_3} = 1.2$ respectively. $v_2$ should be embedded closer to $v_3$ than $v_1$ in the embedding space because the weight between $v_2$ and $v_3$ is higher than the weight between $v_2$ and $v_1$. Similarly, if we have two nodes $v_4$ and $v_5$ connected by a bidirectional edge and $v_5$ and $v_6$ are connected in one direction, $v_5$ is embedded closer to $v_4$ than $v_6$ as they are connected in both directions. Although directed and weighted edges can provide more structural information to build an effective embedding, these properties are not well exploited in undirected and un-weighted homogeneous graph. That's why researchers start to explore the weighted and/or directed graph and an important number of works (Cao, Lu, & Xu, 2016; Jin et al., 2016; L. Liu, Cheung, Li, & Liao, 2016; Jian Tang et al., 2015; Tian, Gao, Cui, Chen, & Liu, 2014) have been emerged to preserve weights and directions of edges in the embedding space.
*Heterogeneous graph:* in which nodes and edges are of different types. For example, a citation network can be modeled as heterogeneous network where authors, papers and venue represent different types of nodes and relationships between these nodes types (author, paper or venue) represent edges. Preserving nodes and edges types respectively is an interesting and challenging problem and therefore researchers have began to be aware of the importance of heterogeneous network embedding. Community-based question answering, multimedia network and knowledge graphs (Cai, Zheng, & Chang, 2018) represent the main three scenarios where heterogeneous network embedding occurs (Chang et al., 2015; Geng, Zhang, Bian, & Chua, 2015; Ochi et al., 2016). The ultimate goal of heterogeneous network embedding is to map the multimodal objects in the heterogeneous network to a common space such that the intrinsic structures of the original graph is preserved and the similarity between objects can be efficiently computed. For instance, Chang et al. (2015) proposed a deep embedding algorithm for heterogeneous networked data. They considered different types of nodes (images, videos and text documents) and edges that can take place between nodes and they used the proposed embedding algorithm to transform the heterogeneous network into *low-dimensional* space.

*Graph with auxiliary information:* many types of auxiliary information may come with graphs; labels, attribute, node feature, etc. This auxiliary information is crucial for network inference since it preserves more semantic information in the network. A label is a categorical value of a node or an edge which belongs to an information class. Nodes having different labels should be embedded far away from each other. To match node embedding for measuring graph similarity, Nikolentzos et al. (2017) exploit node and edge labels when calculating graph kernels. In more sophisticated way, Xie et al. (2016) proposed an embedding approach in which a more complicated knowledge graph was used. The entity categories are organized in a hierarchical structure, i.e., for the "*book*" category two sub-categories, "*author*" and "*written_work*", are defined. As a second type of auxiliary information, an attribute value is a discrete or continuous value. A significant number of works are developed to embed a graph with discrete node attribute value (Wei, Xu, Cao, & Yu, 2017), continuous high-dimensional node attribute value (Dai, Dai, & Song, 2016) and with both discrete and continuous attribute values for nodes and edges (Niepert, Ahmed, & Kutzkov, 2016). Node feature is also among auxiliary information that can be provided in a graph. Most of node features take textual format provided either as a feature vector for each node or as a document for the whole graph. Image features (Chang et al., 2015) are also possible as another node feature type. Moreover, node features provide rich information available in many real-world networks to enhance the performance of network embedding and to make it inductive (Z. Yang, Cohen, & Salakhutdinov, 2016). Other auxiliary information includes knowledge base (Z. Yang, Tang, & Cohen, 2015), user-location (Alharbi & Zhang, 2016) and information propagation (C. Li, Ma, Guo, & Mei, 2017) and it is not limited to one type. In brief, the importance of the auxiliary information as complementary information sources is even more substantial when the network structure is relatively sparse. Methodologically, the challenge of embedding a graph with auxiliary information is how to combine the topological and auxiliary information in the embedding space. Yang et al. (2015) and Natarajan and Dhillon (2014) proposed multimodal and multisource fusion explored techniques in this line of research.

### 7.2. Context graph: definition and representation

A context graph is an auxiliary neighborhood graph, built from the original network $G$, in which similar nodes share an edge between them and for any *source* node $v$ is associated its one-hop neighbor as its *context*. A context vertex of another vertex can be in the immediate neighborhood or in the k-hop neighborhood as mentioned in figure 3. To define the context graph, researchers proposed a variety of methods that can be categorized as follows: random-walk based methods, adjacency matrix based methods, and miscellaneous (personalized PageRank based methods and direct matrix based methods). In fact, the simplest way to define a context graph is the use of the adjacency matrix of a given network $G$ and in which two nodes sharing a link between them are similar to each other. Other methods exploit the higher order neighborhood via random walks to measure similarity among nodes.

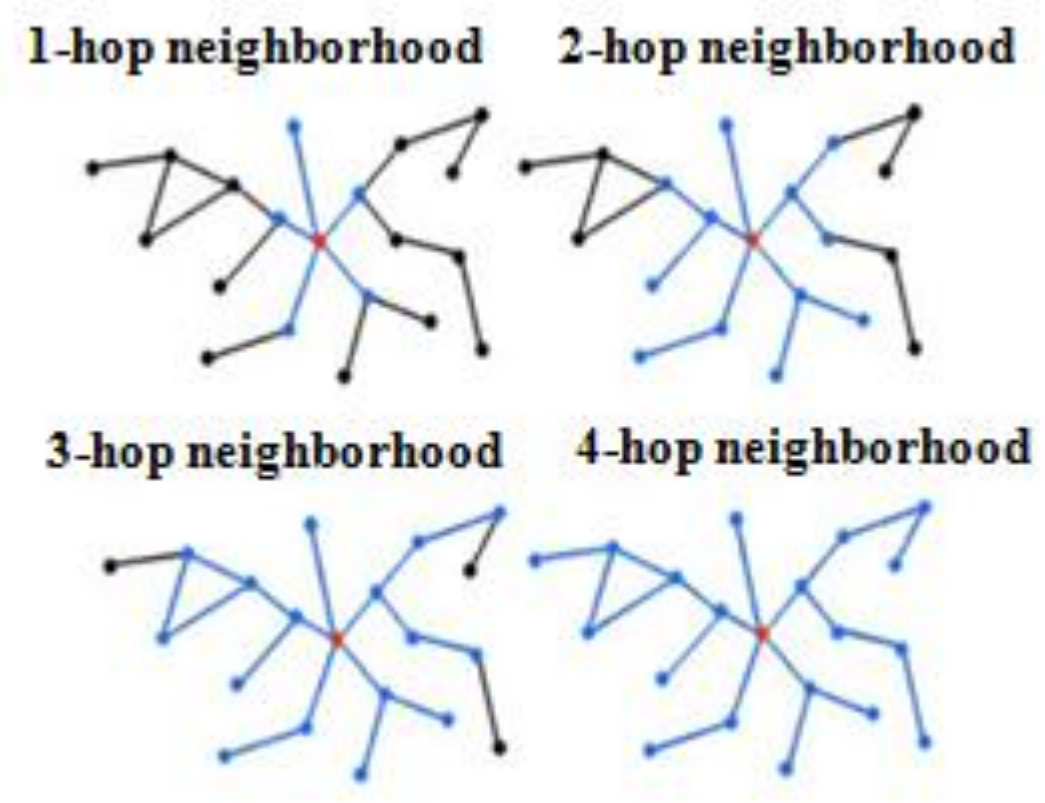

**Figure 3:** k-hop neighborhood

7.2.1. Random-walk based methods

One of the most challenges of network embedding is to preserve network structure, i.e., the higher order neighborhood structure, in the embedding space. In random-walk based methods, to define the context graph, a set of random-walk paths sampled from the original network $G$ have been used to approximate graph properties by capturing global and local structural information. In fact, global structural information can be preserved by the high-order proximity and local structural information is preserved by means of first-order proximity. As the name suggests, a random-walk selects, at first, a staring node $v$ and then randomly transiting through the edges until reaching a maximum length $l$. The length of a random-walk is manually fixed before as well as the number of random walks $\gamma$ to start from each vertex.

***Definition 3***: Formally, a random-walk is defined as follows: Given a graph $G = (V, E)$, a random walk is a sequence of vertices $v_1, v_2 \ldots \ldots v_l$ where $v_i \in V$ with $1 \leq i \leq l$ and $(v_i, v_{i+1}) \in E$ for all $1 \leq i \leq l - 1$ such that $l$ represents the length of the random walk.

A random-walk of length $l$ starts at node $v_i$ in the graph $G$, then migrates to a neighbor vertex $v_j$ of $v_i$, and then migrates to the neighbor of $v_j$ and so on until reaching $l$ steps. The choice of the starting node as well as the transition node in each step is either randomly uniform or based on some distribution. In the former, the transition node is picked randomly based on a stochastic process. In the latter, a probability distribution is defined to control the choice of the transition node and this choice is no longer random but it is biased and the decision of where to walk next is influenced by some distribution probability. In the view of these insights, two popular models that fall into this category are DeepWalk (Perozzi et al., 2014) and node2vec (Grover & Leskovec, 2016). DeepWalk is the pioneer research work in using random walks to learn nodes representations. Indeed, it performs truncated random walks by uniformly sampling a neighbor of the last visited node until reaching a maximum length of the walk, to generate nodes sequences. Authors in deepWalk defined a random walk as "a stochastic process with random variables $W_{v_i}^1; W_{v_i}^2; : : : ; W_{v_i}^k$; such that $W_{v_i}^{k+1}$ is a vertex chosen at random from the neighbors of vertex $v_k$" and they denoted random-walk rooted at vertex $v_i$ as $W_{v_i}$. In particular, for each $v_k \in W_{v_i}$ and for each $v_j \in W_{v_i}[k-r:k+r]$ (r is the window size), $(v_k, v_j)$ forms an edge in the corresponding context graph.

Unlike deepWalk that performs a uniform random-walk, node2vec uses a second order random-walk which means that the choice of the next node depends on the two last visited nodes. To provide more flexibility in exploring node's neighborhood, node2vec further exploits a biased random walks that provide a trade-off between breadth-first (BFS) and depth-first (DFS) graph searches, to capture the local and global information respectively, by introducing two hyperparameters, $p$ and $q$. The hyperparameter $p$, i.e., return parameter, has been used to control the likelihood of the walk immediately revisiting a node. In other words, $p$ controls the probability to go back to the previous node after visiting the current node. The hyperparameter $q$, i.e., in-out parameter, is used to control the likelihood of the walk revisiting a node's one-hop neighborhood.

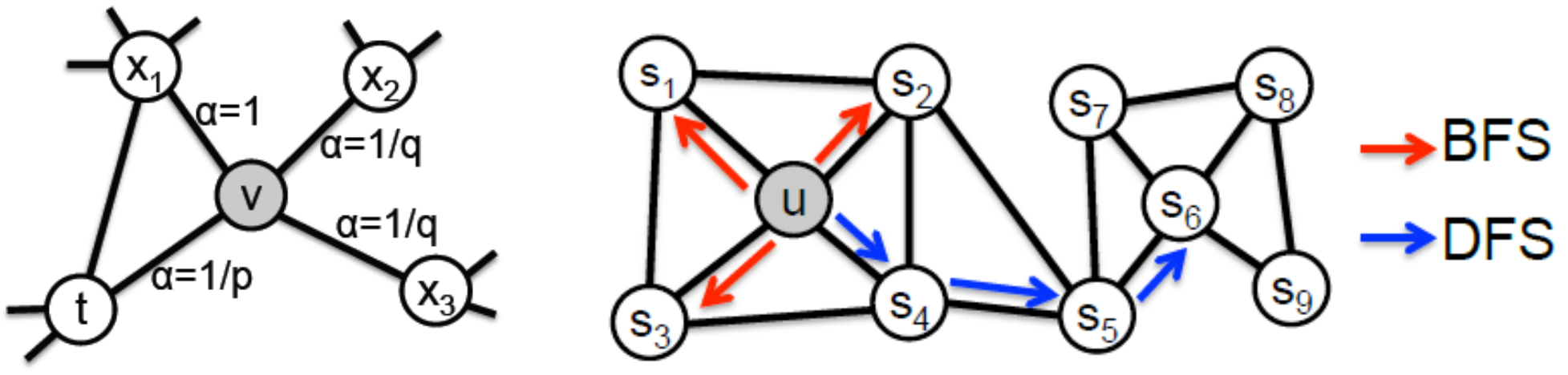

**A** **B**

**Figure 4: A.** biased random walk procedure of node2vec. **B.** BFS and DFS search strategies from node u (Grover & Leskovec, 2016)

For instance, in figure 4, if the walk just went from node $t$ to node $v$, it must select an adjacent node $x$ of $v$ to visit (it can be $t$ again). The probability of transition from node $v$ to node $x$ relied by an edge weighted by $w_{vx}$ is defined by:

$$P(c_i = x | c_{i-1} = v) = \begin{cases} \frac{\pi_{vx}}{Z} & if (v,x) \in E \\ 0 & otherwise \end{cases} \quad (3)$$

where $\pi_{vx}$ is the unnormalized transition probability between nodes $v$ and $x$, $Z$ is the normalizing constant and $c_i$ and $c_{i-1}$ denote respectively the *i-th* and *(i-1)-th* nodes in the walk.

As the random walk now resides at node *v* (figure 3), it needs to decide on the next step so it evaluates the transition probabilities $\pi_{vx}$ on edges (*v*, *x*) leading from *v*. Authors set the unnormalized transition probability to $\pi_{vx}$= $\alpha_{pq}(t,x). w_{vx}$ where:

$$\alpha_{pq}(t,x) = \begin{cases} \frac{1}{p}, & if\ d_{tx} = 0 \\ 1, & if\ d_{tx} = 1 \\ \frac{1}{q} & if\ d_{tx} = 2 \end{cases} \quad (4)$$

and $d_{tx}$ denotes the shortest path distance between nodes *t* and *x* such that $d_{tx}$ must be one of {0 ,1 ,2} and hence, the two parameters $p$ and $q$ are necessary and sufficient to guide the walk. Consequently, the trade-off between BFS and DFS strategies in node2vec allows capturing more flexible contextual structure better than deepWalk. As figure 3.B implies, BFS strategy is beneficial for learning local neighbors and DFS is better for learning global structure of the network.

### 7.2.2. Adjacency based methods

Adjacency based methods directly exploit the given graph as its context graph. Nodes that share either an edge between them or share the common immediate neighbors are embedded closer in the embedding space. Examples of such methods are Large-scale Information Network Embedding (LINE) (Jian Tang et al., 2015) and Structural Deep Network Embedding (SDNE) (D. Wang et al., 2016a). LINE can preserve the first- and second-order proximities.

***Definition 4:*** The first-order proximity is defined as the local pairwise proximity between two connected nodes. For each pair of nodes linked by an edge $(v_i, v_j)$, its weight $w_{ij}$ designates the first-order proximity between $v_i$ and $v_j$. If there is no edge between two nodes, the first-order proximity is 0. In figure 5, the observed link between nodes 6 and 7 represents the first-order proximity between them.

***Definition 5:*** The second-order proximity between two nodes $v_i$ and $v_j$ is determined by the similarity of $v_i$ and $v_j$'s neighborhood structures. In figure 5, to obtain the second-order proximity between the two nodes 5 and 6, the similarity between their neighborhoods 1, 2, 3 and 4 have to be measured.

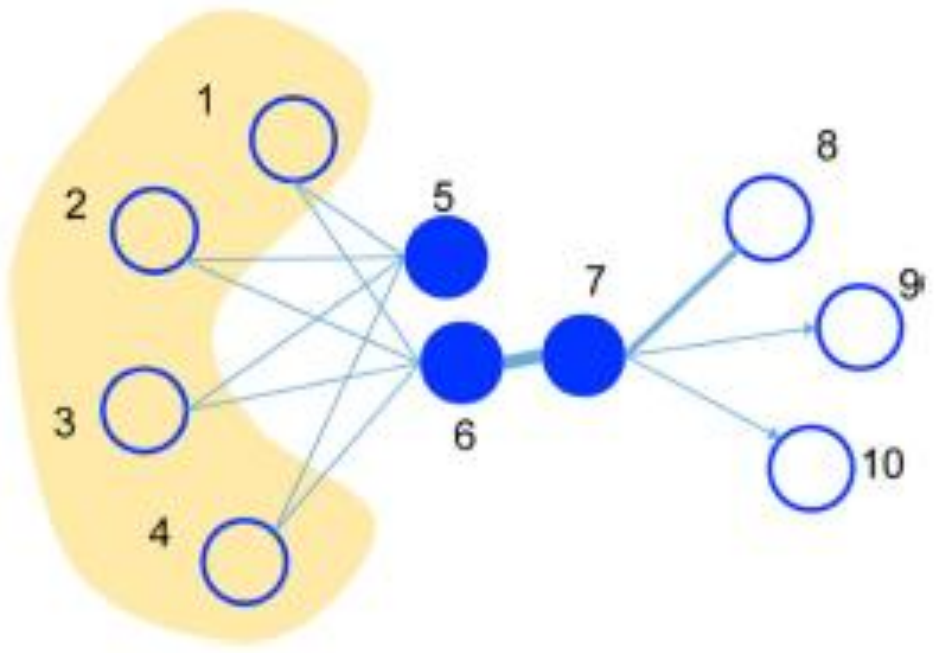

**Figure 5:** Example of first and second-order proximities (Jian Tang et al., 2015)

These two proximities are crucial to measure the relationships between two nodes. Accordingly, two joint probability distributions are defined to quantify such measures. The first-order proximity between two nodes $v_i$ and $v_j$ is measured by the following joint probability distribution:

$$P_1(v_i, v_j) = \frac{1}{1 + \exp(-\vec{u}_i^T \vec{u}_j)} \qquad (5)$$

where $\vec{u}_i \in R^d$ and $\vec{u}_j \in R^d$ represent the low-dimensional representations of nodes $v_i$ and $v_j$ respectively.

The second-order proximity can be quantified by the probability of the context node $v_j$ being generated by the node $v_i$ as follows:

$$P_2(v_j | v_i) = \frac{\exp(\vec{u}'^T_j \vec{u}_i)}{\sum_{k=1}^{|V|} \exp(\vec{u}'^T_k \vec{u}_i)} \qquad (6)$$

where $|V|$ is the number of nodes, $\vec{u}_i$ is the representation of $v_i$ when it is treated as a vertex and $\vec{u}'_i$ is the representation of $v_i$ when it is treated as a context node.
This conditional distribution implies that nodes with similar distributions over the contexts are similar to each other.

GraphSAGE (W. Hamilton, Ying, & Leskovec, 2017) is also among adjacency based method used to define context graph. It is characterized by a parameter $k$ to control the neighborhood depth. In figure 6, the parameter $k$ is 2 and nodes at distance 2 are seen in the same neighborhood. After defining the neighborhood, an information sharing procedure has to be defined between neighbors. Aggregators are defined for such use. The aggregator functions take a neighborhood as input and combine each neighbor's embedding with weights to form a neighborhood embedding. Weights of the aggregator are either learned or fixed depending on the used function. Formally, GraphSAGE exploits a two layer deep neural architecture and in each layer $k$ a node $v$ from the graph $G$ calculates its representation $h^k$ which represents an aggregation of representations of its neighbors from the previous layer, $\{h^{k-1}(u), \forall\, u \in N(v)\}$ where $N(v)$ represents the neighbors of $v$.

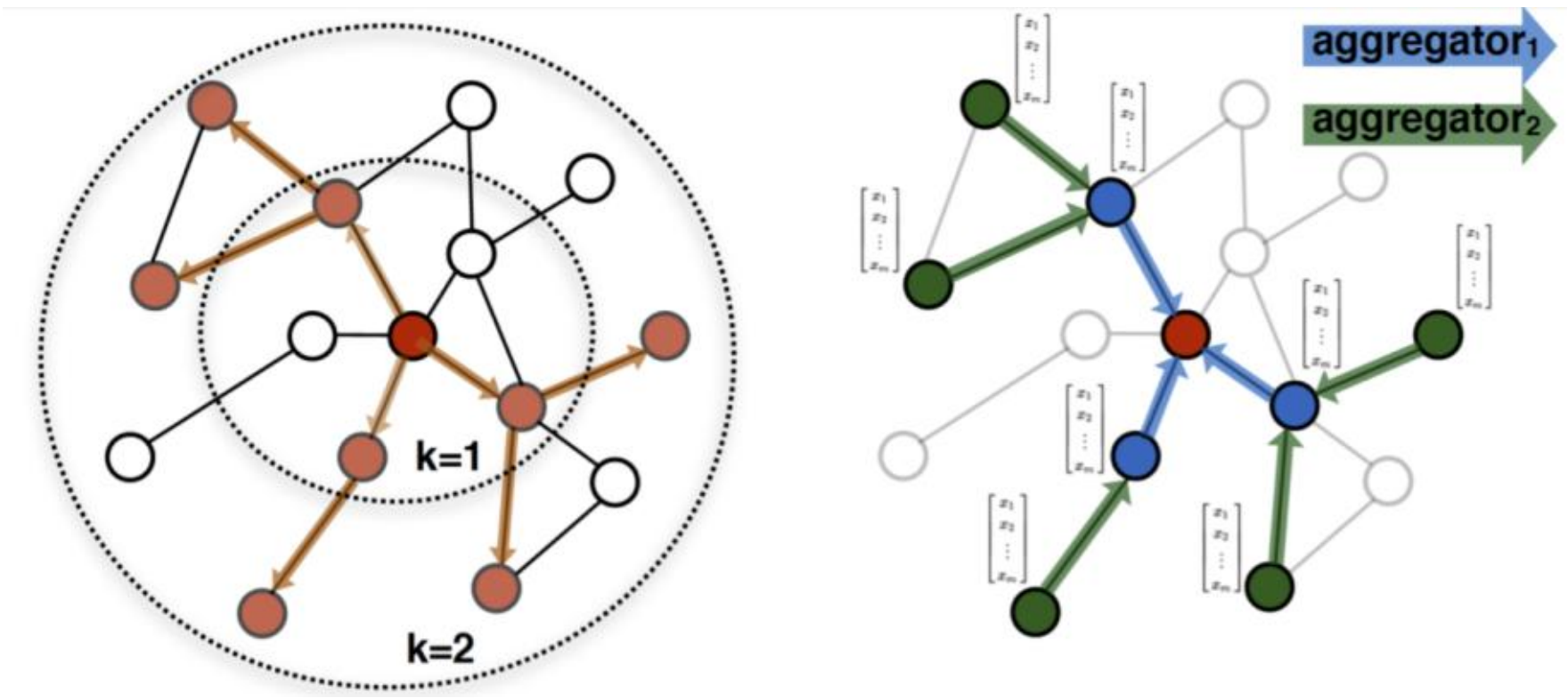

**Figure 6:** Neighborhood exploration and information aggregation in GraphSAGE (W. Hamilton et al., 2017)

### 7.2.3. Miscellaneous

There are other proposed approaches used to generate a context graph. Ou et al. (2016) uses similarity notions to quantify similarity among nodes and define the similarity matrix as the context matrix and then they use it to learn embeddings preserving the asymmetric transitivity of nodes in directed network. Asymmetric transitivity indicates that if there is an edge from node $v_i$ to node $v_j$ and another edge from node $v_j$ to node $v_k$, there is likely an edge from node $v_i$ to node $v_k$ but not in the opposite direction. To measure such property, the similarity matrix is constructed based on four similarity measures, summarized in a general formulation: Katz similarity[11], common neighborhoods, Rooted PageRank[12], Adamic-Adar[13]. Tsitsulin et al. (2018) use the Personalized PageRank[14] to measure similarity between nodes of the graph and exploits the similarity matrix as context matrix.

[11] Katz similarity represents a centrality measure to quantify influence of a given node within a social network. To measure influence, it uses a weighted summation over the path set between two nodes.

[12] Rooted PageRank (RPR) is an adaption of the PageRank algorithm for the link prediction task. The RPR between two participants x and y is defined by the stationary probability distribution of y under the following random walk:

- With probability $\alpha$, jump to node x
- With probability 1- $\alpha$, jump to a random neighbor of the current node

[13] Adamic-Adar is a variant of common neighbors' measure and it assigns each neighbor a weight which is the reciprocal of the degree of the neighbor.

[14] Personalized PageRank (PPR) is a PageRank variation often used for recommender systems and it is biased towards a set of source nodes.

### 7.3. Latent network representation learning models

After building the context graph, different models can be adopted to transform the original network representation into the embedding space. The most commonly used models consist of matrix factorization[15] based models, skip-gram based models, deep learning based methods, and other methods.

#### 7.3.1. Matrix factorization

As shown above, some context graph approaches use the adjacency matrix to define the context graph. The matrix values indicate the relationship among nodes in the original network so that each row or column of the matrix is used as the vector representation of a given node with $|V|$ dimensions, such that $|V|$ is the total number of nodes in the network. In this sense, it is an intuitively way to learn node embeddings using matrix factorization. Singular value decomposition (SVD) is one the commonly used matrix factorization model in network embedding due to its optimality for low-rank decomposition (Ou et al., 2016). High-Order Proximity preserved Embedding (HOPE) (Ou et al., 2016), Collaborative Representation Learning (CRL) (Junyang Chen, Gong, Wang, Liu, & Dong, 2021), and GraRep (Cao, Lu, & Xu, 2015), a novel model for learning graph representations for knowledge management, are recent proposed approaches that fall firmly within this class of network embedding based on matrix factorization. Effectively, HOPE applied a generalized SVD to obtain low-dimensional representations of nodes and it preserves the higher-order proximity by minimizing the following loss of approximation:

$$\min \left\| S - U^s {U^t}^T \right\| \qquad (7)$$

where $S$ is the similarity matrix which is used to define the context graph as mentioned in section 7.2.3 , $U^s, U^t \in R^{|V| \times d}$ are the source embedding matrix and the target embedding matrix respectively.

#### 7.3.2. Skip-gram based models

These models are inspired from the popular NLP Word2vec model. By analogy to the Word2vec algorithm, after performing random walks to build paths, each random-walk corresponds to a sentence from the corpus and each node in the random-walk is considered as word in these sentences. Noting that the distribution of nodes in short random walks is similar to the distribution of words in natural language, skip-gram model, a widely used Word2vec model, is adopted by DeepWalk to learn node embeddings. Given the following walk $s = \{v_1, v_2, .., v_l\}$ of length $l$ , DeepWalk aims to maximize the probability of the neighbors of node $v_i$ in this walk conditioned on its embedding $\emptyset(v_i)$ as follows:

$$\max_{\emptyset} \mathrm{P}(\{v_{i-c}, .., v_{i+c}\} \backslash v_i | \emptyset(v_i)) = \prod_{j=i-c, j \neq i}^{i+c} \mathrm{P}(v_j | \emptyset(v_i)) \quad (8)$$

where $\{v_{i-c}, .., v_{i+c}\} \backslash v_i$ is the local context nodes of node $v_i$, $c$ defines the window size and $\emptyset(v_i)$ is the current embedding of node $v_i$.

[15] Matrix factorization is a way of decomposing the original matrix into two rectangular matrices with low-dimensions.

DeepWalk uses also hierarchical softmax (Mikolov et al., 2013) to efficiently infer latent representations that preserve the higher order node proximity.

Node2vec is another example of skip-gram based representation learning models. In the same way as deepWalk, node2vec aims to preserve higher-order proximity between nodes and it demonstrated that deepWalk is not expressive enough to capture the diversity of connectivity patterns in a network. Walk paths in deepWalk are sampled based on depth-first sampling (DFS) strategy and neighboring nodes composing these paths are sampled at increasing distance from the source node sequentially. Thus, a few nodes close to node source will be sampled and the local structure may be neglected. To preserve the local structure, the breadth-first sampling (BFS) strategy is exploited. It explores immediate nodes to the source node. That's why node2vec leverages a biased random-walk strategy that interpolates between BFS and DFS strategies. Sampled paths are then fed into the skip-gram model to learn node representations in such way that nodes with the same network community or sharing similar roles are closely embedded. Similar to deepWalk, node2vec uses Negative Sampling (Mikolov et al., 2013) to efficiently infer node embeddings. Negative sampling is a technique used to train machine learning models which present more negative observations compared to positive ones. For each training sample, a small percentage of the model's weights are updated rather than updating all of them and hence reducing the computational complexity of the model. The main idea behind Negative Sampling concept is to boost a given vertex to be close to its neighbors and in the meanwhile be far from its negative samples. These negative sampling are chosen based on the frequencies of vertices without considering if that these samples are really negative to the target vertex. To address this problem, some recent attempts are proposed for improving Negative Sampling. Chen et al (2021) present a recent method, called Hierarchical Negative Sampling (HNS), used to model the latent structures of vertices and learn the relations among them and draw more appropriate negative samples and thereby obtain better performance on network embeddings.

#### 7.3.3. Deep learning based models

The main goal of network embedding is to map nodes from the high-dimensional network into a low-dimensional space. In fact, network data are highly non-linear and thus requiring an effective non-linear function to map the original network to an embedding space. Deep models are certainly the most adequate techniques for such use because of their success in other fields. Some deep learning approaches (Cao et al., 2016; D. Wang et al., 2016a; S. Wang, Tang, Aggarwal, Chang, & Liu, 2017) have been proposed to address such problems and to impose network structure and property-level constraints on deep models. Wang et al. (2016a) proposed SDNE, an approach based on deep autoencoders[16] to address non-linearity in network structure and preserve first and second-order proximities. SDNE (D. Wang et al., 2016a) exploits an autoencoder with multiple non-linear layers and simultaneously optimizes the first and second-order proximity to obtain the embeddings. The model is composed of two parts: unsupervised and supervised. The former is composed of an autoencoder used to learn embedding, which preserves the second-order proximity, for a node. The latter is used to learn embedding, which preserves the first-order proximity, for a node by means of Laplacian Eigenmaps[17] (LE) (Belkin & Niyogi, 2003). Figure 7 shows the framework of SDNE where $K$ represents the number of hidden layers and $\{y_i^k\}_{i=1}^{|V|}$represents the *k-th* layer hidden representations.

The autoencoder aims to minimize the $L_2$ norm[18] ($\|.\|_2$) as the reconstruction error between the input and the output by the following loss function:

$$\tau = \sum_1^{|V|} \|\hat{x}_i - x_i\|_2^2 \qquad (9)$$

where $|V|$ is the total number of vertices in the network, $\hat{x}_i$ represents the reconstructed vector (output vector) , and $\boldsymbol{x_i}$ denotes the input vector.

Another example of deep models applied to learn nodes embeddings is GraphSAGE. GraphSAGE uses an aggregation method based on deep learning architecture composed of two layers to learn embeddings in inductive way which means that the prediction of an embedding for a new node does not require retraining the whole graph. It learns a set of aggregator functions that integrate features of local neighborhood and pass it to the target node to induce the embedding of a new node.

[16] An autoencoder is an unsupervised technique for neural networks used to compress and encode data and then learn how to reconstruct data back from the reduced representation to a representation closer to the original input as much as possible.
[17] Laplacian Eigenmaps is a linear model for low-dimensional linear projection that preserves distances for all points.
[18] $L_2$ norm is defined as a standard method to calculate the length of a vector in Euclidian space.

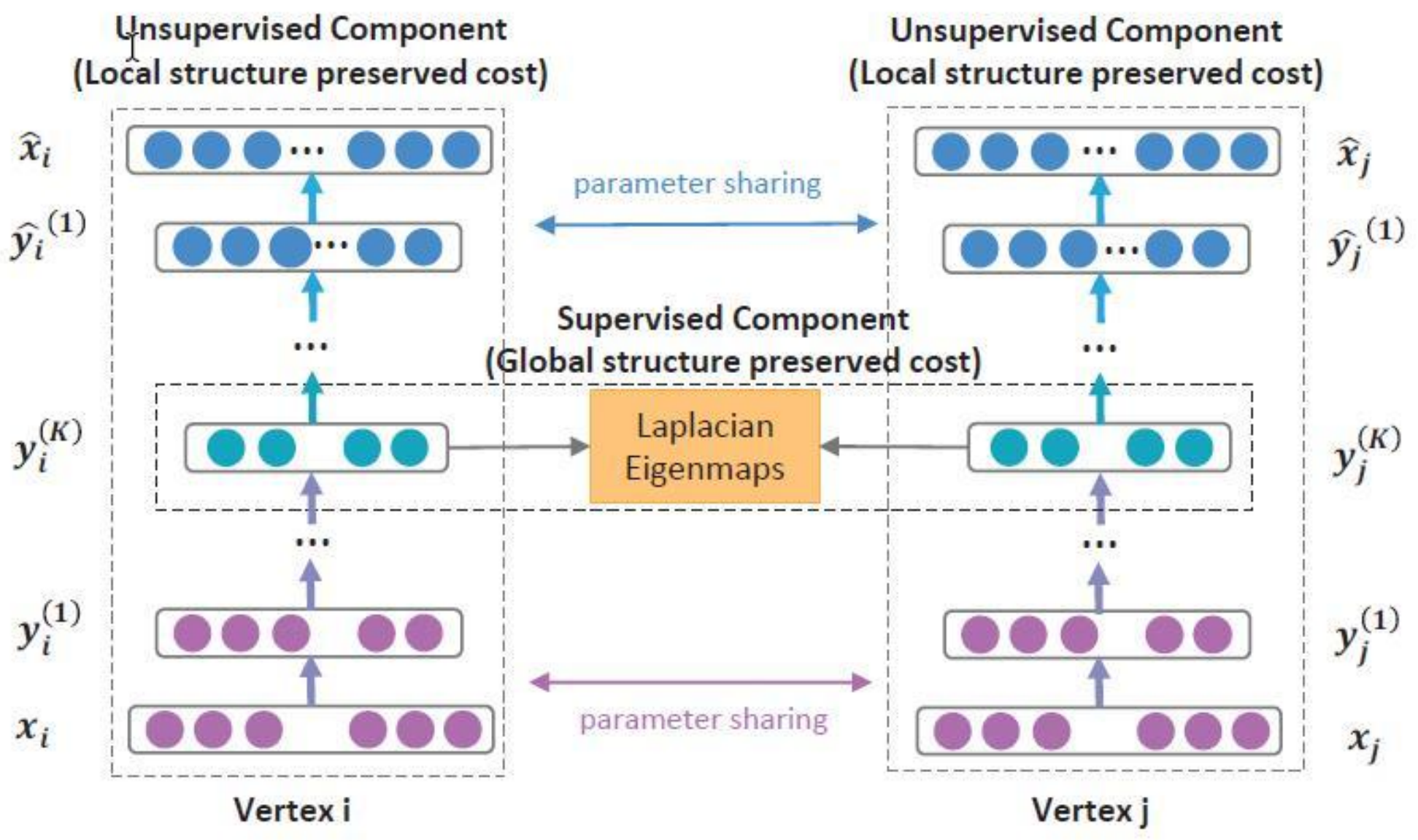

**Figure 7:** The framework of SDNE (D. Wang et al., 2016a)

Furthermore, one of the most recent advancement in deep learning are graph attention networks which are based on an attention mechanism dealing with variable sized inputs and focusing on the most relevant parts of the input to make decisions. Accordingly, graph attention networks have demonstrated its potential in network representation learning. Wang et al. (2019) proposed HAN (Heterogeneous graph Attention Network) framework which uses a two-level attention mechanism. HAN takes into consideration both of node-level and semantic-level attention mechanisms and utilizes meta-paths to model higher-order proximity. The representation of node $u$ is aggregated from its meta-path based neighbors and the two-level attention mechanism is used to learn the weights of neighbors.

### 7.3.4. Other methods

As denoted above, LINE aims to preserve the first and second-order proximity and it defines two joint probability distributions[19] to quantify these two similarities. The first-order proximity is modeled above by the equation (3) over the space $V \times V$ and in order to learn node embeddings the empirical probability[20] between $v_i$ and $v_j$ is defined below by equation (8):

$$\hat{P}_1(v_i, v_j) = \frac{w_{ij}}{\sum_{(i,j)\in E} w_{ij}} \qquad (10)$$

[19] A joint probability is defined as the probability of two events occurring together.

[20] Empirical probability is the probability of an event based on the results of an actual experiment conducted various times.

where $w_{ij}$ represents the weight of the link between the two nodes $v_i$ and $v_j$, and $E$ represents the set of edges in a given network.

Then, LINE minimizes the combination of the two distributions $P_1$ and $\hat{P}_1$ by means of Kullback-Leibler[21] (KL) divergence as a loss function as follows:

$$O_1 = KL(\hat{P}_1, P_1) = -\sum_{(i,j)\in E} w_{ij} \log\left(P_1(v_i, v_j)\right) \qquad (11)$$

Similarly, to preserve second-order proximity, authors define probability distributions and a loss function.

7.4. Network embedding output

Network embedding provides as output a (or a set of) low-dimensional vector(s) for a given network. Different types of output can be produced depending on what part of the graph we would to embed and what type of application will use the output embedding. For that reason, network embedding output is divided into 4 categories: node embedding, edge embedding, hybrid embedding, and whole graph embedding.

*Node embedding:* representing a node as a vector can benefit a wide variety of node related network analysis applications. Multi-label node classification, node recommendation, and node clustering are among these applications that can be performed efficiently in terms of time and space when using such vector representations. Nonetheless, network analysis applications take profit from networked data not only on the node level but also in the edge or sub-graph levels and even on the whole graph level.

Most of the network representation learning models presented above provide a node embedding as output in which each node in the network is represented by a low-dimensional vector in the embedding space. The main difference between these network representation learning models resides on how they define "*closeness*" between two vertices based on which they will be embedded closer in the embedding space. Similarity measure between a pair of nodes can be computed through various metrics such as higher-order proximity, first-order proximity, second-order proximity, and so forth. In some works, structural identity, i.e., a concept of symmetry, is also leveraged to measure similarity between two nodes. For instance, Figueiredo et al. (2017) proposed *struc2vec*, a novel representation learning framework for capturing structural identity of nodes in a network in such way that nodes having similar structural roles are characterized by close embeddings in the embedding space as shown in figure 8.

[21]The KL divergence is a non-symmetric measure of the difference between two probability distributions *p(x)* and *q(x)*.

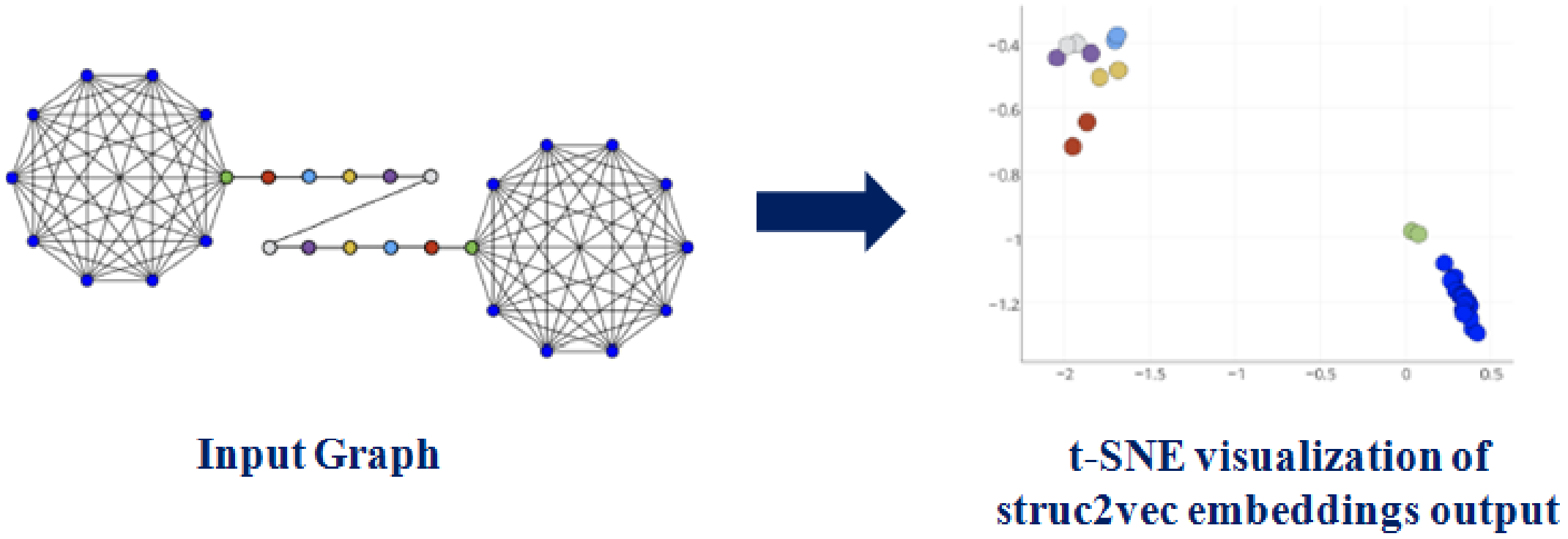

**Figure 8:** Structural identity preserving with struc2vec (Figueiredo et al., 2017)

*Edge embedding:* it maps each edge in a given network into a low-dimensional vector. Knowledge graphs, link prediction, and community detection are among domain applications of edge embedding. In knowledge graphs, edge embeddings are used to preserve relations between each head and tail entities pair in the latent space (Bordes, Usunier, Garcia-Duran, Weston, & Yakhnenko, 2013; Bordes, Weston, Collobert, & Bengio, 2011). For example, Bordes et al. (2013) present a new approach called *TransE* to learn relationships in knowledge graphs. In *TransE*, relationships are represented as translations in the latent space: if the triplet (head entity, relation, tail entity) exists then the embeddings of head and tail entities should be embedded closer depending on the relationship between the entity pair. In case of link prediction task, an edge embedding can be fed into a classifier to predict whether an edge is likely to exist or not. Friend recommendation as a classical link prediction problem gradually leverages edge embedding to discover links that can take place in the future in a given social network. In this sense, Verma et al. (2019) proposed an approach based on a neural network architecture to learn edge embedding for friend recommendation. The approach takes as input a collection of user-pairs and then learns to predict new edges. Regarding community detection as an application of edge embeddings, Junyang Chen, Gong, Dai, Yuan, & Liu (2020) proposed an adversarial learning approach for modeling overlapping communities of vertices in order to map each community and vertex into an embedding space while learning the association between each pair of them.

*Hybrid embedding:* different components of a network, i.e., node and edge or node and community, can be combined and learned together. The combination of node and edge embeddings is also known as *substructure embedding* which has gained an important attention from the research community. Yanardag and Vishwanathan (2015) and Narayanan et al. (2016) learn a vector representation of subgraphs, graphlets[22] and rooted sub-graphs[23] respectively, using Word2vec architecture where each subgraph represents a word in the corpus. Another research work (Adhikari, Zhang, Ramakrishnan, & Prakash, 2018) focuses on the embedding of sub-graphs having arbitrary structure. Similar to substructure embedding, community embedding has also attracted researchers' attention. Cavallari et al. (2017) present a novel Community Embedding framework *ComE* which jointly solves community embedding, community detection and node embedding together and they define community embedding as a multivariate Gaussian distribution[24] which is used in turn to empower community detection from the node embedding results by a Gaussian mixture[25] formulation.

[22]Graphlets refer to a sub-structure type used to decompose graphs and they are induced, non-isomorphic sub-graphs of a large network.

*Whole graph embedding:* in where a graph will be represented by a low-dimensional vector such that similar graphs are embedded closer to each other in the embedding space. Gene regulation networks, protein networks and molecule networks are examples in which the whole-graph embedding takes place. Such embedding is beneficial for graph classification task which requires an efficient solution for computing graph level similarity. Li et al. (2017) present *DeepCas*, a novel deep learning based approach that learns the representation for a whole cascade graph[26] in an end-to-end manner which refers to the fact that all parameters are trained jointly in the intention of improving the performance of cascade prediction[27].

[23] A rooted sub-graph is a sub-graph in which a node is distinguished as the root.

[24] Multivariate Gaussian distribution is defined as a generalization of the one dimensional normal distribution to a vector of multiple normally distributed variables.

[25] Gaussian mixture is a function consisting of several Gaussians.

[26] A cascade graph refers to a set of cascade paths sampled through multiple random walk processes.

[27] Cascade prediction refers to the fact of identifying when a piece of information goes viral in social media.

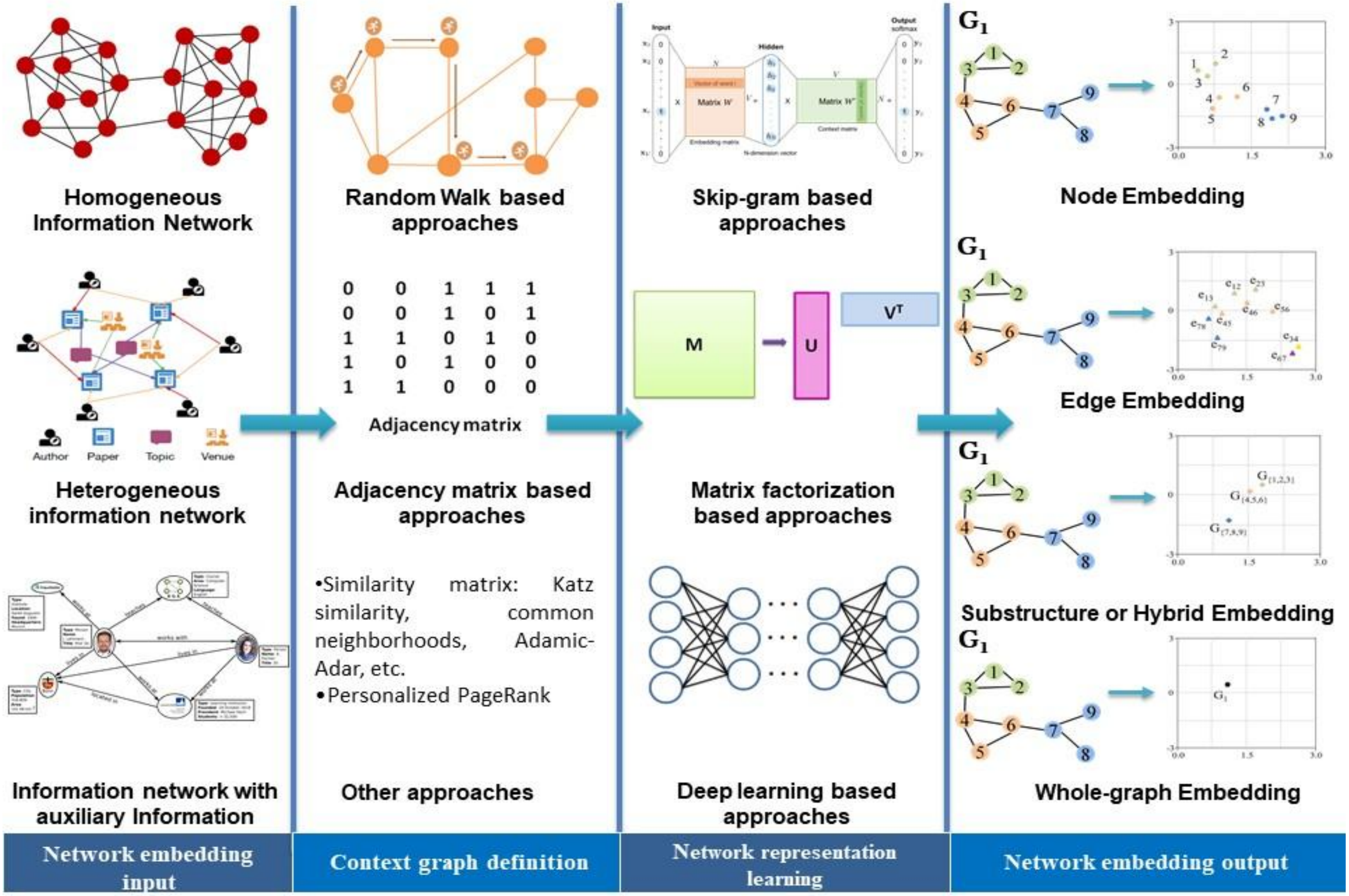

**Figure 9:** Categorization of different settings and approaches used in each step of network embedding pipeline

## 8. Downstream tasks

Once embeddings are learned through network representation learning techniques, a wide variety of network analytics applications, known also as downstream tasks, can profit from such embeddings as the vector representation can be efficiently processed in both time and space. In this section, we introduce, in a more comprehensive way, the most popular downstream tasks or applications of network embedding – Link Prediction (LP), Node Classification (NC), Network Visualization (NV), and Recommendation systems (RS). For each downstream task, four dimensions are highlighted: main idea, network representation learning contribution, used datasets and evaluation metrics. Moreover, extra downstream tasks are also highlighted such as knowledge graph and social networks alignment. Table 5 categorizes NRL models regarding the used downstream task and network type used for evaluation.

**Table 3: Classification of NRL models**

| | Information Network input type | | | Context graph definition | | | Network embedding Technique | | | | Output of NRL model | | | |
|---|---|---|---|---|---|---|---|---|---|---|---|---|---|---|
| **Model** | Homogeneous | Heterogeneous | Graph with auxiliary information | Random walk based | Adjacency based | Miscellaneous | Skip-gram based | Deep learning based | Matrix factorization based | Others | Node embedding | Edge embedding | Hybrid Embedding | Whole graph embedding |
| **DeepWalk** | ✓ | _ | _ | ✓ | _ | _ | ✓ | _ | _ | _ | ✓ | _ | _ | _ |
| **GraphSAGE** | ✓ | _ | ✓ | _ | ✓ | _ | _ | ✓ | _ | _ | ✓ | _ | _ | _ |
| **GraRep** | ✓ | _ | _ | _ | ✓ | _ | _ | _ | ✓ | _ | ✓ | _ | _ | _ |
| **HOPE** | ✓ | _ | _ | _ | _ | ✓ | _ | _ | ✓ | _ | ✓ | _ | _ | _ |
| **Sub2vec** | ✓ | _ | _ | ✓ | _ | _ | ✓ | _ | _ | _ | _ | _ | ✓ | _ |
| **LINE** | ✓ | _ | _ | _ | ✓ | _ | _ | _ | _ | ✓ | ✓ | _ | _ | _ |
| **Node2vec** | ✓ | _ | _ | ✓ | _ | _ | ✓ | _ | _ | _ | ✓ | ✓ | _ | _ |
| **SDNE** | ✓ | _ | _ | _ | ✓ | _ | _ | ✓ | _ | _ | ✓ | _ | _ | _ |
| **DeepCas** | ✓ | _ | _ | ✓ | _ | _ | _ | ✓ | _ | _ | _ | _ | _ | ✓ |
| **ComE** | ✓ | _ | _ | ✓ | _ | _ | _ | _ | ✓ | _ | ✓ | _ | ✓ | _ |
| **struc2vec** | ✓ | _ | _ | ✓ | _ | _ | ✓ | _ | _ | _ | ✓ | _ | _ | _ |
| **HIN2Vec** | _ | ✓ | ✓ | ✓ | _ | _ | _ | ✓ | _ | _ | ✓ | ✓ | _ | _ |
| **HAN** | _ | ✓ | ✓ | _ | ✓ | _ | _ | ✓ | _ | _ | ✓ | _ | _ | _ |

## 8.1. Link Prediction

LP is defined as the task of predicting either missing interactions in a graph or links that may appear in the future in an evolving network. LP is one of the most fundamental problems of network analysis and it is more precisely at the core of recommender systems. For instance, friendship links between two users who know each other are missing and they can be predicted using such systems. Furthermore, many biological interaction graphs, e.g., protein-protein interaction network, are incomplete and the verification of the existence of links between nodes requires a costly experimental tests. Hence, LP methods are applied in biological network analysis (Martınez, Cano, & Blanco, 2014) with the attention of reducing the cost of empirical approaches.

Most attempts on network embedding driven link prediction are focusing on homogeneous networks (Grover & Leskovec, 2016; Perozzi et al., 2014; Jian Tang et al., 2015). Node2vec (Grover & Leskovec, 2016) learns node embedding for three types of networks: Facebook, protein-protein interaction (PPI) network and the collaborative ARXIV network. The Facebook network is composed of one type of nodes (user node) and one type of relationship (friendship relation) between nodes. In the PPI network, nodes are proteins and relationships between nodes indicate a biological interaction among proteins. The collaborative network is composed of scientists who represent nodes and a collaboration relationship between nodes if the two nodes have collaborated in a paper. Learned embeddings of these homogeneous networks are then used to predict missing edges between nodes such as missing friendship relation and biological interaction. Relatively fewer work (Fu, Lee, & Lei, 2017; Zhuo, Zhan, Liu, Xie, & Lu, 2019) on network embedding copes with heterogeneous network link prediction. Fu et al. (2017) presented a novel representation learning framework called *HIN2Vec* (Heterogeneous Information Network to Vector) for heterogeneous networks and evaluated the efficiency of the learned embeddings through link prediction task. HIN2Vec is able to capture various types of links among nodes and thus preserving very rich information. A *meta-path* defined by a sequence of nodes types and/or nodes edges is usually exploited to denote a particular relationship between a pair of nodes so that the different semantics of relationships are better captured. Learned vectors by HIN2Vec are used to predict missing links between different types of nodes.

- *Representation learning*: In network embedding, learned low-dimensional vectors should be able to preserve different levels of network proximity, such as first-order proximity, second-order proximity, structural equivalence, and so on, and thus capturing explicit and implicit connections between nodes of the network to enable application of link prediction. In HOPE (Ou et al., 2016), authors used learned representations which preserve high-order proximity to predict missing edges in Twitter and the social network Tensent Weibo[28], a Twitter-style social platform in China. In addition, Grover and Leskovec (2016) demonstrated that the predicted links using embeddings, that preserve both homophily[29] and structural roles of nodes in the network, are more accurate than links predicted by traditional similarity based LP approaches.
- *Evaluation metrics*: The most common used evaluation metrics (Jinyin Chen et al., 2019; Lü & Zhou, 2011) to measure the efficiency of LP task are as follows: Precision@$k$, Mean Average Precision@$k$ (MAP), F1-score, accuracy, and Area Under receiver operating Curve (AUC). Their definitions are given below:
    - Precision@k is defined by the percentage of top-k evaluated pairwise nodes that are existing edges of the ground truth. It is defined as follows:

      $$Precision@k = \frac{\sum_{i=1}^{k} \mu_i}{k} \qquad (12)$$

      where $k$ is the number of evaluated pairwise nodes, $\mu_i$ is a binary variable indicating whatever the $i$-th predicted pair of nodes is correct ($\mu_i = 1$) or not ($\mu_i = 0$).
    - Mean Average Precision@k (MAP) is the mean of the average scores for the ranking results for each vertex $v$. The average precision (AvP) is computed as the sum of top-k precision (precision@k). AvP and MAP equations are given below:

      $$AvP(v) = \sum_{k=1}^{n} \frac{Precision@k}{k} \qquad (13)$$

      $$MAP = \frac{\sum_{i=1}^{K} AvP(v_i)}{K} \qquad (14)$$

      where $n$ represents the size of the edges set.
    - Area Under receiver operating Curve is defined as the probability that the predictor gives a higher score to a randomly chosen existing link than a randomly chosen non-existing link. It is calculated as follows:

[28] http://t.qq.com/

[29] Homophily refers to the fact that nodes are organized based on communities they belong to.

$$AUC = \frac{n' + 0.5\, n''}{n} \qquad (15)$$

where $n$ is the number of times of randomly picking a pair of links from missing links set and unconnected links set, $n'$ is the number of times that the missing link is having a higher score than unconnected link, while $n''$ is the number of times when they have the same score.

To illustrate, node2vec used the AUC evaluation metric to measure the efficiency of LP task using embeddings. SDNE approach (D. Wang et al., 2016a) used precision@*k* as the evaluation metric to predict hidden links in sparse networks using learned vectors. All of these works show that low-dimensional vectors can infer missing links with high precision.

- *Datasets*: For conducting LP using learned representations researchers exploit a diversity of real-world networks which are organized into three categories: online social networks, citations networks and biological networks. SN-TWeibo[30], SN-Twitter (De Choudhury et al., 2010), Facebook (Leskovec & Krevl, 2014) are among online social networks used datasets. For citation networks, there are three popular used datasets in LP: ARXIV[31] (Leskovec, Kleinberg, & Faloutsos, 2007; Leskovec & Krevl, 2014), DBLP[32] (Jie Tang et al., 2008) and CORA[33] (McCallum, Nigam, Rennie, & Seymore, 2000). Regarding the biological networks, the most used dataset is Protein-Protein Interaction network (PPI)[34] (Breitkreutz et al., 2007), the biological network representing the pairwise physical interactions between proteins in yeast. For example, node2vec has used the three categories of datasets (Facebook dataset, PPI and ARXIV datasets) to evaluate LP task using learned vectors.

## 8.2. Node Classification

Node classification (NC) refers to the prediction of the class label for each unlabeled node in a graph by training a set of labeled nodes. Labels, in a social network, may indicate interests, affiliations, or beliefs. While in a citation network labels may indicate the topic or the research area of the publication, they show the functionality of a protein in a PPI network. Depending on the number of labels to be affected to each node, the task is classified into single label and multi-label node classification (Kong, Shi, & Yu, 2011). In social networks, for instance, users choose to make their data either private or public and they do not provide all information in their accounts such as demographic information. Missing labels can be inferred through labeled nodes and links of the network.

[30] http://www.kddcup2012.org/c/kddcup2012-track1/data
[31]http://snap.stanford.edu/data/ca-AstroPh.html
[32]http://arnetminer.org/citation
[33] https://linqs.soe.ucsc.edu/node/236
[34] http://konect.uni-koblenz.de/networks/maayan-vidal

- *Representation learning:* NC is the most used downstream task to evaluate the efficiency of network embedding outputs or learned representations. Due to their capabilities to capture the intrinsic structure of the original network, learned representations have demonstrated its efficiency in node classification. Indeed, learned vectors of the labeled nodes are used to train a classification model. Then, the learned vectors of unlabeled nodes are fed into the learned classification model to infer their labels. Many recent works including deepWalk (Perozzi et al., 2014), node2vec (Grover & Leskovec, 2016), SDNE (D. Wang et al., 2016a), HOPE (Ou et al., 2016), and LINE (Jian Tang et al., 2015) have evaluated the predictive power of network embedding under various types of networks such as social, biological and collaborative networks. These works have demonstrated that better vector representations can contribute to high classification accuracy. DeepWalk leverages learned representations to predict labels in three social networks: BlogCatalog, Flickr and YouTube. For BlogCatalog, embeddings are fed into a one-vs-rest logistic regression model for classification to assign topic categories to authors. Similarly, classification model is used to predict interest groups for users and groups of viewers, which enjoy common video genres, in Flickr and YouTube social networks respectively.
- *Evaluation metrics*: The common used metrics (Khosla et al., 2019) to evaluate the efficiency of the classification task are mirco-F1 and macro-F1. These two metrics are defined as follows:
    - Marco-F1 is a measure that gives an equal weight to each class. It is computed by the following formula:

$$Macro - F1 = \frac{\sum_{A \in C} F1(A)}{|C|} \qquad (16)$$

where $A$ is a label, $C$ represents the overall label set and $F1(A)$ is the F1-score for the label $A$.

- Micro-F1 is a metric which gives equal weight to each instance. It is defined as follows:

$$Pr = \frac{\sum_{A \in C} TP(A)}{\sum_{A \in C}(TP(A) + FP(A))} \qquad (17)$$

$$R = \frac{\sum_{A \in C} TP(A)}{\sum_{A \in C}(TP(A) + FN(A))} \qquad (18)$$

$$Micro - F1 = \frac{2 * Pr * R}{Pr + R} \qquad (19)$$

where $TP(A)$, $FP(A)$, and $FN(A)$ represent the number of true positives, false positives and false negatives in the instances which are predicted as A, respectively.

In particular, Perozzi et al. (2014) used the macro-F1 and micro-F1 metrics and reported the average of performance in terms of both to measure the efficiency of deepWalk in multi-label classification task. LINE and GraphSAGE both of them used similarly the two metrics to evaluate the effectiveness of the produced embeddings.

- *Datasets*: Network embedding approaches for node classification is evaluated with various information networks: social networks, citation networks, biological networks, language networks. For social networks, there are three popular used datasets which are Flickr (L. Tang & Liu, 2009), YouTube (Cui et al., 2018), and BlogCalalog (Reza & Huan, 2009). For citation networks, DBLP (Jie Tang et al., 2008), CORA (McCallum et al., 2000), and Citeseer (McCallum et al., 2000) are among the most commonly used datasets for NC. PPI (Grover & Leskovec, 2016) and Wikipedia (Mahoney, 2011) represent the examples of biological and language networks respectively. As an illustration, BlogCatalog and Flickr are used in deepWalk, node2vec and SDNE to test the performance of embeddings with classification task and they showed a promising classification results. Moreover, Citeseer and CORA are used in a variety of works (Pan, Wu, Zhu, Zhang, & Wang, 2016; Tu, Zhang, Liu, Sun, & others, 2016; Cheng Yang et al., 2015). Wikipedia as information network has also been studied in a wide range of recent works (Grover & Leskovec, 2016; Jian Tang et al., 2015; Tu et al., 2016). To evaluate the quality of word embeddings, Tang et al. (2015) used the words vector representations to calculate the document representation which can be evaluated by a document classification task to classify document into 7 diverse categories: history, arts, human, mathematics, nature, technology, and sports.

Table 4 summarizes the above downstream tasks, datasets and their evaluation metrics.

**Table 4:** Downstream tasks, used datasets, and evaluation metrics

| Downstream Task | Datasets | Evaluation metrics |
|---|---|---|
| Link Prediction | • Online social networks: SN-TWeibo, SN-Twitter (De Choudhury et al., 2010), Facebook (Leskovec & Krevl, 2014)<br>• Citation networks: ARXIV (Leskovec et al., 2007; Leskovec & Krevl, 2014), DBLP (Jie Tang et al., 2008) and CORA (McCallum et al., 2000),<br>• Biological networks: Protein-Protein Interaction network (PPI) (Breitkreutz et al., 2007) | Precision @*k*, Mean Average Precision @*k* (MAP), F1-score, accuracy, and Area Under receiver operating Curve (AUC) |
| Node Classification | • Online social networks: Flickr (L. Tang & Liu, 2009), YouTube (Cui et al., 2018), BlogCalalog (Reza & Huan, 2009)<br>• Biological networks (PPI (Grover & Leskovec, 2016)<br>• language networks: Wikipedia (Mahoney, 2011)<br>• Citation networks: DBLP (Jie Tang et al., 2008), CORA (McCallum et al., 2000), and Citeseer (McCallum et al., 2000) | Mirco-F1 and Macro-F1 |

## 8.3. Network visualization

Visualization of complex networked data is powerful for data analysis and exploration. After obtaining low-dimensional representations of nodes, these vectors are passed to a network visualization tool to generate a meaningful visualization that layouts the network in two-dimensional space. The most commonly used visualization tools are *t-distributed Stochastic Neighborhood Embedding* (t-SNE) (Maaten & Hinton, 2008) and *Principal Components Analysis* (PCA). Each embedding algorithm aims to preserve some properties of a graph so that the visualization of the same dataset may differ across several embedding methods. Figure 10 shows the visualization of 20-NewsGroup dataset embeddings using five different embedding algorithms: SDNE, LINE, deepWalk, GraRep and LE. SDNE applied t-SNE tool to obtain clusters of documents by topics.

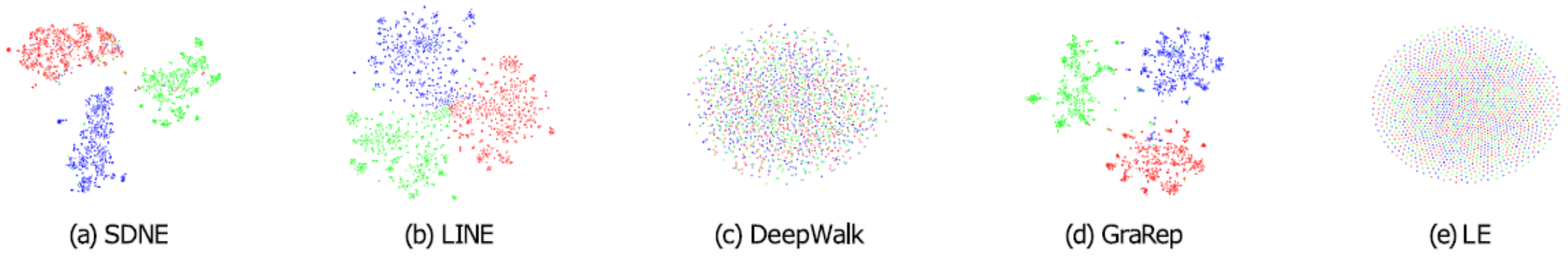

**Figure 10:** Network visualization of 20-NewsGroup dataset by various embedding algorithms using t-SNE (D. Wang, Cui, & Zhu, 2016b)

In figure 10, each node in the network represents a document and different colors correspond to labels. In addition, DBLP co-authorship network is visualized by the LINE (Jian Tang et al., 2015) embedding algorithm to group together authors in the same domain.

## 8.4. Recommendation Systems

Recommendation Systems (RS) have been playing an important role in various online services since their main purpose is to recommend suitable items for interested users given the explosive growth of information on the web. Traditionally, there are two widely used approaches for recommendation systems: Collaborative Filtering (CF) and Content based Filtering (CBF). CF is based on the basic assumption that users sharing similar preferences will consume similar items. As the name suggests, CBF is a technique used to recommend items similar to items that have previously interested in a specific user. Although these recommendation systems' models have achieved a great success in the literature, they suffer from some drawbacks such as cold-start problem and sparse ratings problem with limited information. To address these drawbacks, network embedding is integrated for recommendation systems to help in producing quality recommendations and improve recommendation model's accuracy. The following sub-sections present some research works focusing on exploiting network embedding for recommendation systems, used evaluation metrics, and popular datasets for such downstream task.

- *Representation learning:* It worth pointing out that network embedding has been applied to recommendation systems (Ali, Kefalas, et al., 2020; S. Zhang, Yao, Sun, & Tay, 2019) in a new light. In this direction, many research works (Ali, Qi, Muhammad, Ali, & Abro, 2020; Shi, Hu, Zhao, & Philip, 2018; C. Zhang et al., 2017; Zhao et al., 2020; C. Zhou, Liu, Liu, Liu, & Gao, 2017) have been emerged to learn low-dimensional latent vectors for generating different entities recommendations in different online services such as user-item recommendation (Shi et al., 2018), user-location recommendation (C. Zhang et al., 2017; C. Zhou et al., 2017), textual content for paper recommendation (Ali, Qi, et al., 2020). Zhao et al. (2020) proposed a heterogeneous co-occurrence network to learn the latent representations of users and items and then they integrate the two latent representations into one single low-dimensional vector to make recommendations. Ali et al. (2020) proposed a heterogeneous network embedding model that jointly learns embedding for papers and users employing the relationships between different weighted graphs to generate paper recommendations.
- *Evaluation metrics:* There are plenty of metrics (Ali, Kefalas, et al., 2020) used to evaluate the efficiency of recommender systems such as precision, recall, F-measure, Accuracy, and so forth. For instance, Ali et al. (2020) present the most popular metrics used in the literature for evaluating the efficiency of citation recommender systems. Christoforidis et al. (2021) present a unified model that jointly learns user's and points-of-interest (POI) dynamics while providing personalized POI recommendations. They evaluated their proposed model using the accuracy metric. Xie et al. (2021) propose a meta-graph embedding method for item recommendation by leveraging deep neural networks and the attention mechanism. They adopted Mean Absolute Error (MAE) and Root Mean Squared Error (RMSE) for performance evaluation. On the contrary, Ali et al. (2020) employed three other metrics such as recall, Mean Average Precision (MAP), and Mean Reciprocal Rank (MRR) to evaluate the performance of their proposed model. It is worth pointing out that more metrics used to evaluate a given model can lead to a better understanding of its impact.

- *Datasets:* Different information networks are used for evaluating the task of recommendation systems such as online social networks and citation networks. In particular, Ali et al. (2020) review different datasets used in the literature for citation recommendation such as the DBLP and ACL anthology as the most used datasets for this category of recommendation systems. For social networks, Foursquare and Gowalla (Christoforidis et al., 2021) are among the most commonly used networks. There are other research works (F. Xie et al., 2021) that use their own self-collected datasets from other online resources to fulfill their model's requirements.

## 8.5. Others applications

In addition to the above mentioned downstream tasks that are commonly discussed in related works, network embedding can benefit other graph analytics applications as the latent representations can be efficiently processed in both time and space. Based on the data existing in the input graph, more specific downstream tasks may survive.

*Knowledge graph related*: Knowledge graph is a new type of data structure in database systems used to encode information structures of billions of heterogeneous objects as well as their heterogeneous relationships. Knowledge graph embedding has demonstrated its effectiveness to predict missing h or t for a relation fact triplet *(h, r, t).* In this respect, Lin et al. (2015) proposed an approach to embed entity and relation in separate entity and relation spaces respectively to predict missing head or tail entities.

*Social networks alignment related*: Social networks alignment is the fact of aligning users across different online social networks. In other words, it refers to predict whether two social accounts in two different online social networks are owned by the same real user. Network embedding is applied in such task in such way that nodes of users of social networks are embedded into a latent space and learned embedding of nodes are used to align users across two different social networks (Lin et al., 2015; Man, Shen, Liu, Jin, & Cheng, 2016).

**Table 5: Classification of NRL models based on used downstream task**

| | Downstream task | | | | | | Network type used for evaluation | | | |
|---|---|---|---|---|---|---|---|---|---|---|
| **Model** | Node classification | Network visualization | Link prediction | Node clustering | Network reconstruction | Node recommendation | Social network | biological network | citation network | language network |
| **DeepWalk** | ✓ | _ | _ | _ | _ | _ | ✓ | _ | _ | _ |
| **GraphSAGE** | ✓ | _ | _ | _ | _ | _ | ✓ | ✓ | ✓ | _ |
| **GraRep** | ✓ | ✓ | _ | ✓ | _ | _ | ✓ | _ | ✓ | ✓ |
| **HOPE** | _ | _ | ✓ | _ | _ | ✓ | ✓ | _ | ✓ | _ |
| **LINE** | ✓ | ✓ | _ | _ | _ | _ | ✓ | _ | ✓ | ✓ |
| **Node2vec** | ✓ | _ | ✓ | _ | _ | _ | ✓ | ✓ | ✓ | ✓ |
| **SDNE** | ✓ | ✓ | ✓ | _ | ✓ | _ | ✓ | _ | ✓ | ✓ |
| **struc2vec** | ✓ | _ | _ | _ | _ | _ | ✓ | ✓ | _ | ✓ |
| **HIN2Vec** | ✓ | _ | ✓ | _ | _ | _ | ✓ | _ | ✓ | _ |
| **HAN** | ✓ | ✓ | _ | ✓ | _ | _ | _ | _ | ✓ | _ |

## 9. Open-source network embedding libraries

There are several open-source implementations of network embedding algorithms, discussed in this paper, provided by their authors in the web-based hosting service, Github for codes. Table 6 provides code links to interested readers to help them in studying further experiments on these algorithms either by exploiting their own datasets or comparing different implementations between each other for evaluation. Moreover, the tremendous development of network embedding approaches has led to the creation of libraries including the implementation of several network embedding algorithms. These libraries are StellarGraph (Data61, 2018), GraphEmbedding[35], GEM (Goyal & Ferrara, 2018, n.d.), and OpenNE[36].

**Table 6:** Github source code of NRL models

| Model | Code source |
|---|---|
| **DeepWalk** | https://github.com/phanein/deepwalk |
| **GraphSAGE** | https://github.com/williamleif/GraphSAGE |
| **GraRep** | https://github.com/ShelsonCao/GraRep |
| **HOPE** | https://github.com/ZW-ZHANG/HOPE |
| **LINE** | https://github.com/tangjianpku/LINE |
| **Node2vec** | https://github.com/aditya-grover/node2vec |
| **SDNE** | http://nrl.thumedia.org/structural-deep-network-embedding |
| **struc2vec** | https://github.com/leoribeiro/struc2vec |
| **HIN2Vec** | https://github.com/csiesheep/hin2vec |
| **HAN** | https://github.com/Jhy1993/HAN |

- StellarGraph is an open source and Python based machine learning library for network analytics. It solves the problem of representation learning for various types of networks including homogeneous, heterogeneous, attributed, and weighted networks. GraphSAGE, Graph ATtension network (GAT), and node2vec are some examples of implemented network representation learning algorithms on StellarGraph for homogeneous networks. HinSAGE, Metapath2Vec, Relational Graph Convolutional Network (RGCN) represent the set of StellarGraph algorithms developed for heterogeneous network representation learning. Furthermore, StellarGraph provides Attri2vec algorithm for representation learning on attributed networks and node2vec algorithm for weighted networks. After learning representation of nodes and/or edges, it is possible with StellarGraph to evaluate these representations in some implemented machine learning downstream tasks such as link prediction and node classification.

[35] https://github.com/shenweichen/GraphEmbedding
[36] https://github.com/thunlp/OpenNE

- GraphEmbedding is also a Python based network representation learning library. This library supports network embedding for homogeneous networks. To learn node representation, GraphEmbedding library implements five algorithms: deepWalk, LINE, node2vec, SDNE, and struc2vec. Although all of these algorithms operate on homogeneous networks, they differ in the graph properties that they preserve when embedding nodes. To illustrate, while SDNE preserves first- and second-order proximities, struc2vec preserves the structural identity of nodes.
- GEM[37] is an open-source Python library providing a unified interface of different network embedding algorithms. It implements a state-of-the-art embedding algorithms including traditional graph embedding algorithms: LLE, LE, Graph Factorization (GF), SDNE, node2vec, and HOPE. Obtained embeddings can be evaluated by GEM through various downstream tasks including link prediction, graph reconstruction, node classification, and visualization.
- OpenNE is an open-source Python framework unifying the input and the output interfaces of different network embedding models and offering scalable options for each model. The implemented models comprise deepWalk, node2vec, LINE, GraRep, TADW (Text-Associated DeepWalk), SDNE, HOPE, Graph Convolutional Network (GCN), and LE. Moreover, learned node representations can be evaluated via a node classification task provided by OpenNE framework. This framework is continuously evolving and its authors affirm that new network embedding models will be implemented according to their released NRL paper[38].

All of these network embedding frameworks are open-source libraries available on GitHub. Figure 11 shows the number of reported issues for each network embedding library on GitHub.

[37] https://github.com/palash1992/GEM
[38] https://github.com/thunlp/nrlpapers

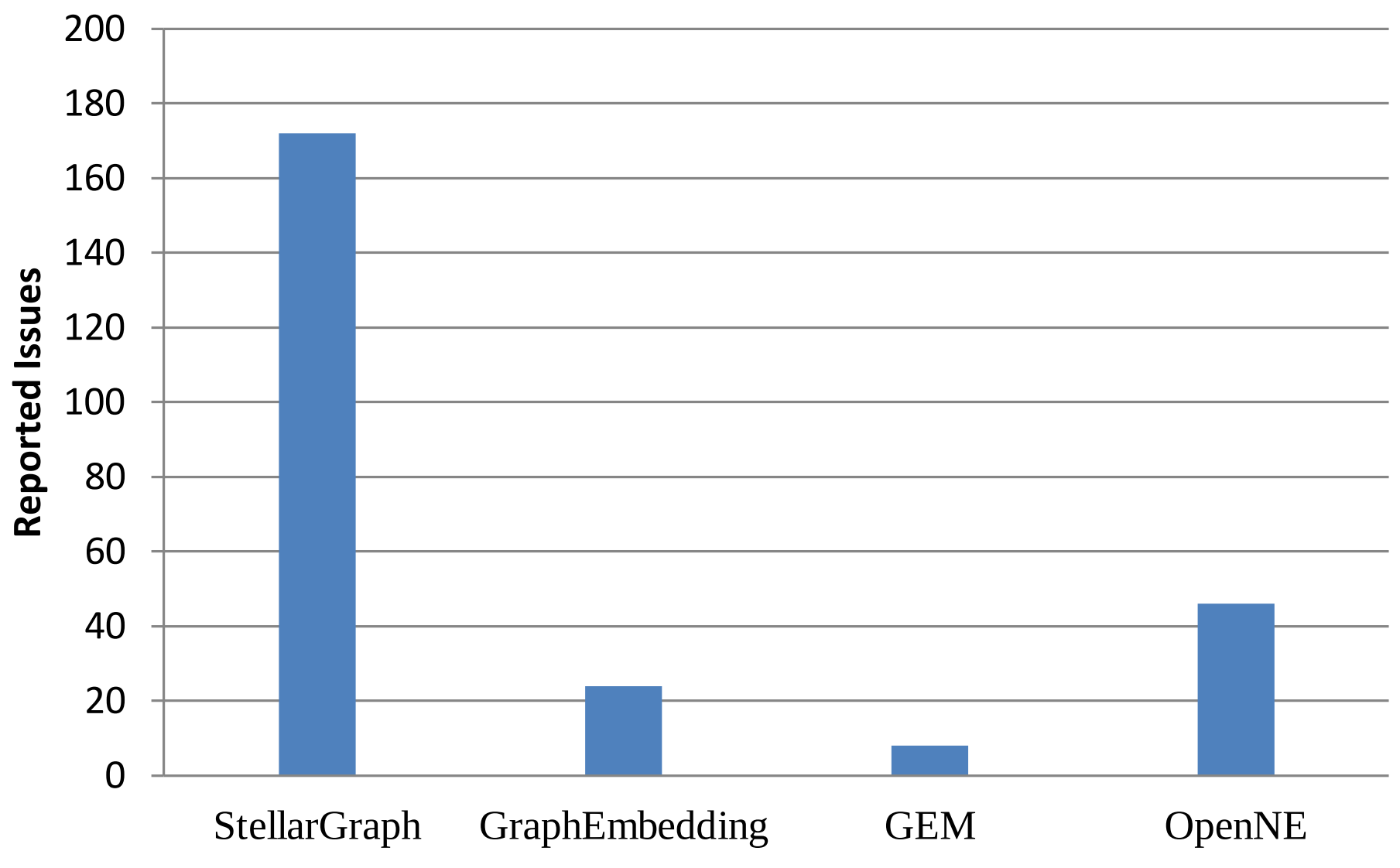

**Figure 11:** Number of reported issues for each network embedding library on GitHub

Network embedding is mainly focusing on large-scale information networks in which the amount of generated data has raised the problem of big data. As a result, there are few attempts to integrate network embedding with big data tools as the case with Neo4j[39], a NoSQL graph database used to store network data. For instance, this graph database provides a plug-in[40] implementing a set of machine learning algorithms that can be applied onto graphs especially the deepWalk representation learning algorithm. DeepWalk can be applied to the whole graph stored in Neo4j and then learns latent representation for each node in the network.

## 10. Conclusion

A comprehensive and systematic survey on network representation learning has been conducted in this paper. We studied the important but unexplored reasons behind network embedding emergence for real-world information networks like online social networks, citation networks, biological networks, etc. In fact, traditional graph embedding techniques focus on graphs with clear gird structure and they are not suitable and applicable for real-world information networks. Therefore, network embedding ancestors ranging from artificial intelligence related context to word representation learning are highlighted in this survey. More importantly, we provide a formal definition of the problem of network embedding along with basic related concepts such as random-walk, first-order proximity, second-order proximity, etc. We further provide the steps composing a network embedding pipeline including network embedding input, context graph definition, network representation learning, and network embedding output. Then, we categorized existing works into different categories for each step of the network embedding pipeline. In the network embedding input step, there exist three types of networks: homogeneous network, heterogeneous network, and network with auxiliary information. In terms of context graph definition, existing approaches are classified as follows: random-walk based, adjacency based, and other types of approaches like similarity based approaches. Similarly, approaches proposed in related works for the third step of network embedding problem are divided into three classes: skip-gram based, matrix factorization based, and deep learning based models. The output of a network embedding problem falls into four categories including node embedding, edge embedding, hybrid embedding, and whole-graph embedding. Learned embeddings should preserve different types of graph properties, such as higher-order proximity and structural identity, depending on the downstream task and hence they can benefit a wide variety of applications. Therefore, we also reported the most common used downstream tasks that network embedding enables, including link prediction, node classification, network visualization, recommendation systems, and other extra applications to evaluate low-dimensional representations. Evaluation of embeddings has demonstrated that better vector representations can contribute to high accuracy as well as high precision in a specific downstream task. Used evaluation metrics and datasets differ from downstream task to another. For instance, link prediction and node classification tasks used different evaluation metrics along with several types of datasets to measure the efficiency of low-dimensional representations. As an important parameter for enhancing this research field, some available open-source network embedding libraries are presented to interested readers to help them in studying further experiments on discussed algorithms.

[39] https://neo4j.com/
[40]https://github.com/neo4j-graph-analytics/ml-models/releases

Although many NRL approaches are proposed in this survey, they are mainly designed for static networks. However, most of real-world networks are evolving over time. For example, in online social networks, relationships between different entities always dynamically change over time. In other words, new edges may be added to the social network and others may be deleted. Consequently, existing network embedding approaches cannot be directly applied to large-scale evolving networks. For that reason, new NRL algorithms are highly desirable to tackle the dynamic aspect of evolving networks. Moreover, most of the existing NRL approaches are designed for general purposes, such as link prediction and node classification, and they mainly focus on general network structures but they may not be specific for some target applications like social ties inferring, rumors detection on social media, and cross-social network information tracking. Due to the complexity of real-world networks, there are more challenging structures and properties that are not considered in the existing NRL approaches. Network motifs and hypernetwork embedding, used to naturally indicate richer relationships among nodes, are among complex structures to be incorporated into an embedding space in an effective manner to provide more consistent and rich embeddings for nodes. Furthermore, as we are living in the era of big data, existing network learning models have to be adapted for big data technologies based on distributed processing to boost NRL models' performance.

## References:

Adhikari, B., Zhang, Y., Ramakrishnan, N., & Prakash, B. A. (2018). Sub2vec: Feature learning for subgraphs. *Pacific-Asia Conference on Knowledge Discovery and Data Mining* (pp. 170–182). Springer. https://doi.org/10.1007/978-3-319-93037-4_14.

Alharbi, B., & Zhang, X. (2016). Learning from your network of friends: A trajectory representation learning model based on online social ties. *2016 IEEE 16th International Conference on Data Mining (ICDM)* (pp. 781–786). IEEE. http://dx.doi.org/10.1109/ICDM.2016.0090.

Ali, Z., Kefalas, P., Muhammad, K., Ali, B., & Imran, M. (2020). Deep learning in citation recommendation models survey. *Expert Systems with Applications*, 113790. Elsevier. https://doi.org/10.1016/j.eswa.2020.113790.

Ali, Z., Qi, G., Muhammad, K., Ali, B., & Abro, W. A. (2020). Paper recommendation based on heterogeneous network embedding. *Knowledge-Based Systems*, *210*, 106438. Elsevier. https://doi.org/10.1016/j.knosys.2020.106438.

Belkin, M., & Niyogi, P. (2003). Laplacian eigenmaps for dimensionality reduction and data representation. *Neural computation*, *15*(6), 1373–1396. MIT Press. http://dx.doi.org/10.1162/089976603321780317.

Bengio, Y., Courville, A., & Vincent, P. (2013). Representation learning: A review and new perspectives. *IEEE transactions on pattern analysis and machine intelligence*, *35*(8), 1798–1828. IEEE. http://dx.doi.org/10.1109/TPAMI.2013.50.

Bengio, Y., Ducharme, R., Vincent, P., & Jauvin, C. (2003). A neural probabilistic language model. *Journal of machine learning research*, *3*(Feb), 1137–1155.

Bordes, A., Usunier, N., Garcia-Duran, A., Weston, J., & Yakhnenko, O. (2013). Translating embeddings for modeling multi-relational data. *Advances in neural information processing systems* (pp. 2787–2795).

Bordes, A., Weston, J., Collobert, R., & Bengio, Y. (2011). Learning structured embeddings of knowledge bases. *Twenty-Fifth AAAI Conference on Artificial Intelligence*.

Breitkreutz, B.-J., Stark, C., Reguly, T., Boucher, L., Breitkreutz, A., Livstone, M., Oughtred, R., et al. (2007). The BioGRID interaction database: 2008 update. *Nucleic acids research*, *36*(suppl_1), D637–D640. Oxford University Press.

Brochier, R., Guille, A., & Velcin, J. (2019). Global vectors for node representations. *The World Wide Web Conference* (pp. 2587–2593). https://doi.org/10.1145/3308558.3313595.

Cai, H., Zheng, V. W., & Chang, K. C.-C. (2018). A comprehensive survey of graph embedding: Problems, techniques, and applications. *IEEE Transactions on Knowledge and Data Engineering*, *30*(9), 1616–1637. IEEE. http://dx.doi.org/10.1109/TKDE.2018.2807452.

Cao, S., Lu, W., & Xu, Q. (2015). Grarep: Learning graph representations with global structural information. *Proceedings of the 24th ACM international on conference on information and knowledge management* (pp. 891–900). https://doi.org/10.1145/2806416.2806512.

Cao, S., Lu, W., & Xu, Q. (2016). Deep neural networks for learning graph representations. *Thirtieth AAAI conference on artificial intelligence*.

Cavallari, S., Zheng, V. W., Cai, H., Chang, K. C.-C., & Cambria, E. (2017). Learning community embedding with community detection and node embedding on graphs. *Proceedings of the 2017 ACM on Conference on Information and Knowledge Management* (pp. 377–386). https://doi.org/10.1145/3132847.3132925.

Chang, S., Han, W., Tang, J., Qi, G.-J., Aggarwal, C. C., & Huang, T. S. (2015). Heterogeneous network embedding via deep architectures. *Proceedings of the 21th ACM SIGKDD International Conference on Knowledge Discovery and Data Mining* (pp. 119–128). https://doi.org/10.1145/2783258.2783296.

Chen, J., Gong, Z., Dai, Q., Yuan, C., & Liu, W. (2020). Adversarial Learning for Overlapping Community Detection and Network Embedding. *ECAI 2020* (pp. 1071–1078). IOS Press.

Chen, J., Gong, Z., Wang, W., & Liu, W. (2021). HNS: Hierarchical negative sampling for network representation learning. *Information Sciences, 542,* 343–356. Elsevier. https://doi.org/10.1016/j.ins.2020.07.015.

Chen, J., Gong, Z., Wang, W., Liu, W., & Dong, X. (2021). CRL: Collaborative Representation Learning by Coordinating Topic Modeling and Network Embeddings. *IEEE Transactions on Neural Networks and Learning Systems*. IEEE.

Chen, J., Zhang, J., Xu, X., Fu, C., Zhang, D., Zhang, Q., & Xuan, Q. (2019). E-lstm-d: A deep learning framework for dynamic network link prediction. *IEEE Transactions on Systems, Man, and Cybernetics: Systems*. IEEE.

Chen, T., & Sun, Y. (2017). Task-guided and path-augmented heterogeneous network embedding for author identification. *Proceedings of the Tenth ACM International Conference on Web Search and Data Mining* (pp. 295–304).

Choi, E., Bahadori, M. T., Searles, E., Coffey, C., Thompson, M., Bost, J., Tejedor-Sojo, J., et al. (2016). Multi-layer representation learning for medical concepts. *Proceedings of the 22nd ACM SIGKDD International Conference on Knowledge Discovery and Data Mining* (pp. 1495–1504).

De Choudhury, M., Lin, Y.-R., Sundaram, H., Candan, K. S., Xie, L., & Kelliher, A. (2010). How does the data sampling strategy impact the discovery of information diffusion in social media? *Fourth International AAAI Conference on Weblogs and Social Media*.

Christoforidis, G., Kefalas, P., Papadopoulos, A. N., & Manolopoulos, Y. (2021). RELINE: point-of-interest recommendations using multiple network embeddings. *Knowledge and Information Systems*, *63*(4), 791–817. Springer. https://doi.org/10.1007/s10115-020-01541-5.

Collobert, R., & Weston, J. (2008). A unified architecture for natural language processing: Deep neural networks with multitask learning. *Proceedings of the 25th international conference on Machine learning* (pp. 160–167).

Cui, P., Wang, X., Pei, J., & Zhu, W. (2018). A survey on network embedding. *IEEE Transactions on Knowledge and Data Engineering*, *31*(5), 833–852. IEEE.

Dahl, G. E., Yu, D., Deng, L., & Acero, A. (2011). Context-dependent pre-trained deep neural networks for large-vocabulary speech recognition. *IEEE Transactions on audio, speech, and language processing*, *20*(1), 30–42. IEEE.

Dai, H., Dai, B., & Song, L. (2016). Discriminative embeddings of latent variable models for structured data. *International conference on machine learning* (pp. 2702–2711).

Data61, C. (2018). StellarGraph Machine Learning Library. *GitHub Repository*. GitHub.

DE, R., Hinton, G., & Williams, R. (1986). Learning internal representations by back-propagating errors. *Nature*, *323*, 533–536.

Eljawad, L., Aljamaeen, R., Alsmadi, M., Almarashdeh, I., Abouelmagd, H., Alsmadi, S., Haddad, F., et al. (2019). Arabic Voice Recognition Using Fuzzy Logic and Neural Network. *ELJAWAD, L., ALJAMAEEN, R., ALSMADI, MK, AL-MARASHDEH, I., ABOUELMAGD, H., ALSMADI, S., HADDAD, F., ALKHASAWNEH, RA, ALZUGHOUL, M. & ALAZZAM, MB*, 651–662.

Elman, J. L. (1990). Finding structure in time. *Cognitive science*, *14*(2), 179–211. Wiley Online Library.

Figueiredo, D. R., Ribeiro, L. F., & Saverese, P. H. (2017). struc2vec: Learning node representations from structural identity. *Proceedings of the 23rd ACM SIGKDD International Conference on Knowledge Discovery and Data Mining, Halifax, NS, Canada* (pp. 13–17).

Fu, T., Lee, W.-C., & Lei, Z. (2017). Hin2vec: Explore meta-paths in heterogeneous information networks for representation learning. *Proceedings of the 2017 ACM on Conference on Information and Knowledge Management* (pp. 1797–1806).

Ganguly, S., Gupta, M., Varma, V., Pudi, V., & others. (2016). Author2vec: Learning author representations by combining content and link information. *Proceedings of the 25th International Conference Companion on World Wide Web* (pp. 49–50). International World Wide Web Conferences Steering Committee.

Geng, X., Zhang, H., Bian, J., & Chua, T.-S. (2015). Learning image and user features for recommendation in social networks. *Proceedings of the IEEE International Conference on Computer Vision* (pp. 4274–4282).

Goyal, P., & Ferrara, E. (2018). Graph embedding techniques, applications, and performance: A survey. *Knowledge-Based Systems*, *151*, 78–94. Elsevier. https://doi.org/10.1016/j.knosys.2018.03.022.

Goyal, P., & Ferrara, E. (n.d.). GEM: A Python package for graph embedding methods. *Journal of Open Source Software*, *3*(29), 876.

Graves, A., Mohamed, A., & Hinton, G. (2013). Speech recognition with deep recurrent neural networks. *2013 IEEE international conference on acoustics, speech and signal processing* (pp. 6645–6649). IEEE.

Grohe, M. (2020). word2vec, node2vec, graph2vec, x2vec: Towards a theory of vector embeddings of structured data. *Proceedings of the 39th ACM SIGMOD-SIGACT-SIGAI Symposium on Principles of Database Systems* (pp. 1–16).

Grover, A., & Leskovec, J. (2016). node2vec: Scalable feature learning for networks. *Proceedings of the 22nd ACM SIGKDD international conference on Knowledge discovery and data mining* (pp. 855–864).

Guo, Y., Liu, Y., Oerlemans, A., Lao, S., Wu, S., & Lew, M. S. (2016). Deep learning for visual understanding: A review. *Neurocomputing*, *187*, 27–48. Elsevier. https://doi.org/10.1016/j.neucom.2015.09.116.

Hamilton, W. L., Ying, R., & Leskovec, J. (2017). Representation learning on graphs: Methods and applications. *arXiv preprint arXiv:1709.05584*.

Hamilton, W., Ying, Z., & Leskovec, J. (2017). Inductive representation learning on large graphs. *Advances in neural information processing systems* (pp. 1024–1034).

Harris, Z. S. (1954). Distributional structure. *Word*, *10*(2-3), 146–162. Taylor & Francis.

Hinton, G. E., & others. (1986). Learning distributed representations of concepts. *Proceedings of the eighth annual conference of the cognitive science society* (Vol. 1, p. 12). Amherst, MA.

Hinton, G. E., & Roweis, S. T. (2003). Stochastic neighbor embedding. *Advances in neural information processing systems* (pp. 857–864).

Hinton, G., McClelland, J., & Rumelhart, D. (1986). Distributed representations. inparallel distributed processing: Explorations in the microstructure of cognition. *Foundations, 1*.

Hussain, S.-A. A., Moosavinasab, S., Sezgin, E., Huang, Y., & Lin, S. (2018). Char2Vec: Learning the Semantic Embedding of Rare and Unseen Words in the Biomedical Literature. *AMIA*.

Jin, Z., Liu, R., Li, Q., Zeng, D. D., Zhan, Y., & Wang, L. (2016). Predicting user's multi-interests with network embedding in health-related topics. *2016 International joint conference on neural networks (IJCNN)* (pp. 2568–2575). IEEE.

Kazemi, S. M., Goel, R., Jain, K., Kobyzev, I., Sethi, A., Forsyth, P., & Poupart, P. (2020). Representation Learning for Dynamic Graphs: A Survey. *Journal of Machine Learning Research*, *21*(70), 1–73.

Khosla, M., Setty, V., & Anand, A. (2019). A Comparative Study for Unsupervised Network Representation Learning. *IEEE Transactions on Knowledge and Data Engineering*. IEEE.

Kong, X., Shi, X., & Yu, P. S. (2011). Multi-label collective classification. *Proceedings of the 2011 SIAM International Conference on Data Mining* (pp. 618–629). SIAM.

Krbec, P. (2006). Language Modeling for Speech Recognition of Czech. Univerzita Karlova, Matematicko-fyzikálnı fakulta.

Krizhevsky, A., Sutskever, I., & Hinton, G. E. (2012). Imagenet classification with deep convolutional neural networks. *Advances in neural information processing systems* (pp. 1097–1105).

Le, Q., & Mikolov, T. (2014). Distributed representations of sentences and documents. *International conference on machine learning* (pp. 1188–1196).

Lee, J. A., & Verleysen, M. (2007). *Nonlinear dimensionality reduction*. Springer Science & Business Media.

Leskovec, J., Kleinberg, J., & Faloutsos, C. (2007). Graph evolution: Densification and shrinking diameters. *ACM transactions on Knowledge Discovery from Data (TKDD)*, *1*(1), 2–es. ACM New York, NY, USA.

Leskovec, J., & Krevl, A. (2014). SNAP Datasets: Stanford large network dataset collection.

Li, B., & Pi, D. (2020). Network representation learning: a systematic literature review. *Neural Computing and Applications*, 1–33. Springer. https://doi.org/10.1007/s00521-020-04908-5.

Li, C., Ma, J., Guo, X., & Mei, Q. (2017). Deepcas: An end-to-end predictor of information cascades. *Proceedings of the 26th international conference on World Wide Web* (pp. 577–586).

Lin, Y., Liu, Z., Sun, M., Liu, Y., & Zhu, X. (2015). Learning entity and relation embeddings for knowledge graph completion. *Twenty-ninth AAAI conference on artificial intelligence*.

Liu, L., Cheung, W. K., Li, X., & Liao, L. (2016). Aligning Users across Social Networks Using Network Embedding. *Ijcai* (pp. 1774–1780).

Liu, Z., Lin, Y., & Sun, M. (2020). Word Representation. *Representation Learning for Natural Language Processing* (pp. 13–41). Springer.

Lü, L., & Zhou, T. (2011). Link prediction in complex networks: A survey. *Physica A: statistical mechanics and its applications*, *390*(6), 1150–1170. Elsevier.

Maaten, L. van der, & Hinton, G. (2008). Visualizing data using t-SNE. *Journal of machine learning research*, *9*(Nov), 2579–2605.

Mahoney, M. (2011). Large text compression benchmark.

Man, T., Shen, H., Liu, S., Jin, X., & Cheng, X. (2016). Predict anchor links across social networks via an embedding approach. *Ijcai* (Vol. 16, pp. 1823–1829).

Martınez, V., Cano, C., & Blanco, A. (2014). ProphNet: a generic prioritization method through propagation of information. *BMC bioinformatics*, *15*(S1), S5. Springer.

McCallum, A. K., Nigam, K., Rennie, J., & Seymore, K. (2000). Automating the construction of internet portals with machine learning. *Information Retrieval*, *3*(2), 127–163. Springer.

Miikkulainen, R., & Dyer, M. G. (1991). Natural language processing with modular PDP networks and distributed lexicon. *Cognitive Science*, *15*(3), 343–399. Wiley Online Library.

Mikolov, T., Chen, K., Corrado, G., & Dean, J. (2013). Efficient estimation of word representations in vector space. *arXiv preprint arXiv:1301.3781*.

Mikolov, T., Kopecky, J., Burget, L., Glembek, O., & others. (2009). Neural network based language models for highly inflective languages. *2009 IEEE International Conference on Acoustics, Speech and Signal Processing* (pp. 4725–4728). IEEE.

Morin, F., & Bengio, Y. (2005). Hierarchical probabilistic neural network language model. *Aistats* (Vol. 5, pp. 246–252). Citeseer.

Mosenthal, P. (1981). Theoretical Issues in Reading Comprehension: Perspectives from Cognitive Psychology, Linguistics, Artificial Intelligence, and Education. JSTOR.

Moyano, L. G. (2017). Learning network representations. *The European Physical Journal Special Topics*, *226*(3), 499–518. Springer.

Narayanan, A., Chandramohan, M., Chen, L., Liu, Y., & Saminathan, S. (2016). subgraph2vec: Learning distributed representations of rooted sub-graphs from large graphs. *arXiv preprint arXiv:1606.08928*.

Natarajan, N., & Dhillon, I. S. (2014). Inductive matrix completion for predicting gene–disease associations. *Bioinformatics*, *30*(12), i60–i68. Oxford University Press.

Niepert, M., Ahmed, M., & Kutzkov, K. (2016). Learning convolutional neural networks for graphs. *International conference on machine learning* (pp. 2014–2023).

Nikolentzos, G., Meladianos, P., & Vazirgiannis, M. (2017). Matching node embeddings for graph similarity. *Thirty-First AAAI Conference on Artificial Intelligence*.

Ochi, M., Nakashio, Y., Yamashita, Y., Sakata, I., Asatani, K., Ruttley, M., & Mori, J. (2016). Representation learning for geospatial areas using large-scale mobility data from smart card. *Proceedings of the 2016 ACM International Joint Conference on Pervasive and Ubiquitous Computing: Adjunct* (pp. 1381–1389).

Ou, M., Cui, P., Pei, J., Zhang, Z., & Zhu, W. (2016). Asymmetric transitivity preserving graph embedding. *Proceedings of the 22nd ACM SIGKDD international conference on Knowledge discovery and data mining* (pp. 1105–1114).

Pan, S., Wu, J., Zhu, X., Zhang, C., & Wang, Y. (2016). Tri-party deep network representation. *Network*, *11*(9), 12.

Perozzi, B., Al-Rfou, R., & Skiena, S. (2014). Deepwalk: Online learning of social representations. *Proceedings of the 20th ACM SIGKDD international conference on Knowledge discovery and data mining* (pp. 701–710).

Pimentel, T., Veloso, A., & Ziviani, N. (2017). Unsupervised and scalable algorithm for learning node representations.

Prusa, J. D., & Khoshgoftaar, T. M. (2016). Designing a better data representation for deep neural networks and text classification. *2016 IEEE 17th International Conference on Information Reuse and Integration (IRI)* (pp. 411–416). IEEE.

Reddy, S., Fox, J., & Purohit, M. P. (2019). Artificial intelligence-enabled healthcare delivery. *Journal of the Royal Society of Medicine*, *112*(1), 22–28. SAGE Publications Sage UK: London, England.

Reza, Z., & Huan, L. (2009). Social computing data repository.

Rosenberg, A. L. (1980). Issues in the study of graph embeddings. *International Workshop on Graph-Theoretic Concepts in Computer Science* (pp. 150–176). Springer.

Roweis, S. T., & Saul, L. K. (2000). Nonlinear dimensionality reduction by locally linear embedding. *science*, *290*(5500), 2323–2326. American Association for the Advancement of Science.

Schmidhuber, J., & Heil, S. (1996). Sequential neural text compression. *IEEE Transactions on Neural Networks*, *7*(1), 142–146. IEEE.

Shi, C., Hu, B., Zhao, W. X., & Philip, S. Y. (2018). Heterogeneous information network embedding for recommendation. *IEEE Transactions on Knowledge and Data Engineering*, *31*(2), 357–370. IEEE.

Socher, R. (2014). *Recursive deep learning for natural language processing and computer vision*. Citeseer.

Tang, J., Qu, M., Wang, M., Zhang, M., Yan, J., & Mei, Q. (2015). Line: Large-scale information network embedding. *Proceedings of the 24th international conference on world wide web* (pp. 1067–1077).

Tang, J., Zhang, J., Yao, L., Li, J., Zhang, L., & Su, Z. (2008). Arnetminer: extraction and mining of academic social networks. *Proceedings of the 14th ACM SIGKDD international conference on Knowledge discovery and data mining* (pp. 990–998).

Tang, L., & Liu, H. (2009). Relational learning via latent social dimensions. *Proceedings of the 15th ACM SIGKDD international conference on Knowledge discovery and data mining* (pp. 817–826).

Tenenbaum, J. B., De Silva, V., & Langford, J. C. (2000). A global geometric framework for nonlinear dimensionality reduction. *science*, *290*(5500), 2319–2323. American Association for the Advancement of Science.

Tian, F., Gao, B., Cui, Q., Chen, E., & Liu, T.-Y. (2014). Learning deep representations for graph clustering. *Twenty-eighth AAAI conference on artificial intelligence*.

Tsitsulin, A., Mottin, D., Karras, P., & Müller, E. (2018). Verse: Versatile graph embeddings from similarity measures. *Proceedings of the 2018 World Wide Web Conference* (pp. 539–548).

Tu, C., Zhang, W., Liu, Z., Sun, M., & others. (2016). Max-margin deepwalk: Discriminative learning of network representation. *IJCAI* (Vol. 2016, pp. 3889–3895).

Uzuner, Ö. (2009). Recognizing obesity and comorbidities in sparse data. *Journal of the American Medical Informatics Association*, *16*(4), 561–570. BMJ Group BMA House, Tavistock Square, London, WC1H 9JR.

Verma, J., Gupta, S., Mukherjee, D., & Chakraborty, T. (2019). Heterogeneous edge embedding for friend recommendation. *European Conference on Information Retrieval* (pp. 172–179). Springer.

Wang, D., Cui, P., & Zhu, W. (2016a). Structural deep network embedding. *Proceedings of the 22nd ACM SIGKDD international conference on Knowledge discovery and data mining* (pp. 1225–1234).

Wang, D., Cui, P., & Zhu, W. (2016b). Structural Deep Network Embedding. *Proceedings of the 22nd ACM SIGKDD International Conference on Knowledge Discovery and Data Mining*, KDD ‑16 (p. 1225‑1234). San Francisco, California, USA: Association for Computing Machinery. Retrieved from https://doi.org/10.1145/2939672.2939753

Wang, Q., Mao, Z., Wang, B., & Guo, L. (2017). Knowledge graph embedding: A survey of approaches and applications. *IEEE Transactions on Knowledge and Data Engineering*, *29*(12), 2724–2743. IEEE.

Wang, S., Tang, J., Aggarwal, C., Chang, Y., & Liu, H. (2017). Signed network embedding in social media. *Proceedings of the 2017 SIAM international conference on data mining* (pp. 327–335). SIAM.

Wang, X., Cui, P., Wang, J., Pei, J., Zhu, W., & Yang, S. (2017). Community preserving network embedding. *Thirty-first AAAI conference on artificial intelligence*.

Wang, X., Ji, H., Shi, C., Wang, B., Ye, Y., Cui, P., & Yu, P. S. (2019). Heterogeneous graph attention network. *The World Wide Web Conference* (pp. 2022–2032).

Wei, X., Xu, L., Cao, B., & Yu, P. S. (2017). Cross view link prediction by learning noise-resilient representation consensus. *Proceedings of the 26th International Conference on World Wide Web* (pp. 1611–1619).

West, J., & Bhattacharya, M. (2016). Intelligent financial fraud detection: a comprehensive review. *Computers & security*, *57*, 47–66. Elsevier.

Wright, J., Ma, Y., Mairal, J., Sapiro, G., Huang, T. S., & Yan, S. (2010). Sparse representation for computer vision and pattern recognition. *Proceedings of the IEEE*, *98*(6), 1031–1044. IEEE.

Xiang, S., Nie, F., Zhang, C., & Zhang, C. (2008). Nonlinear dimensionality reduction with local spline embedding. *IEEE Transactions on Knowledge and Data Engineering*, *21*(9), 1285–1298. IEEE.

Xie, F., Zheng, A., Chen, L., & Zheng, Z. (2021). Attentive Meta-graph Embedding for item Recommendation in heterogeneous information networks. *Knowledge-Based Systems*, *211*, 106524. Elsevier.

Xie, R., Liu, Z., & Sun, M. (2016). Representation Learning of Knowledge Graphs with Hierarchical Types. *IJCAI* (pp. 2965–2971).

Xu, W., & Rudnicky, A. (2000). Can artificial neural networks learn language models? *Sixth international conference on spoken language processing*.

Yanardag, P., & Vishwanathan, S. (2015). Deep graph kernels. *Proceedings of the 21th ACM SIGKDD International Conference on Knowledge Discovery and Data Mining* (pp. 1365–1374).

Yang, C., Liu, Z., Zhao, D., Sun, M., & Chang, E. (2015). Network representation learning with rich text information. *Twenty-Fourth International Joint Conference on Artificial Intelligence*.

Yang, C., Xiao, Y., Zhang, Y., Sun, Y., & Han, J. (2020). Heterogeneous Network Representation Learning: Survey, Benchmark, Evaluation, and Beyond. *arXiv preprint arXiv:2004.00216*.

Yang, Z., Cohen, W. W., & Salakhutdinov, R. (2016). Revisiting semi-supervised learning with graph embeddings. *arXiv preprint arXiv:1603.08861*.

Yang, Z., Tang, J., & Cohen, W. (2015). Multi-modal Bayesian embeddings for learning social knowledge graphs. *arXiv preprint arXiv:1508.00715*.

Zhang, C., Zhang, K., Yuan, Q., Peng, H., Zheng, Y., Hanratty, T., Wang, S., et al. (2017). Regions, periods, activities: Uncovering urban dynamics via cross-modal representation learning. *Proceedings of the 26th International Conference on World Wide Web* (pp. 361–370).

Zhang, D., Yin, J., Zhu, X., & Zhang, C. (2018). Network representation learning: A survey. *IEEE transactions on Big Data*. IEEE.

Zhang, S., Yao, L., Sun, A., & Tay, Y. (2019). Deep learning based recommender system: A survey and new perspectives. *ACM Computing Surveys (CSUR)*, *52*(1), 1–38. ACM New York, NY, USA.

Zhang, Z., Cui, P., & Zhu, W. (2020). Deep learning on graphs: A survey. *IEEE Transactions on Knowledge and Data Engineering*. IEEE.

Zhao, Z., Zhang, X., Zhou, H., Li, C., Gong, M., & Wang, Y. (2020). HetNERec: Heterogeneous network embedding based recommendation. *Knowledge-Based Systems*, *204*, 106218. Elsevier.

Zhou, C., Liu, Y., Liu, X., Liu, Z., & Gao, J. (2017). Scalable graph embedding for asymmetric proximity. *Thirty-First AAAI Conference on Artificial Intelligence*.

Zhou, F., Liu, L., Zhang, K., Trajcevski, G., Wu, J., & Zhong, T. (2018). Deeplink: A deep learning approach for user identity linkage. *IEEE INFOCOM 2018-IEEE Conference on Computer Communications* (pp. 1313–1321). IEEE.

Zhuo, W., Zhan, Q., Liu, Y., Xie, Z., & Lu, J. (2019). Context Attention Heterogeneous Network Embedding. *Computational Intelligence and Neuroscience*, *2019*. Hindawi.

## Appendix 1:

| Model name | Reference |
| --- | --- |
| **act2vec, trace2vec, log2vec, model2vec** | De Koninck, Pieter, Seppe vanden Broucke, and Jochen De Weerdt. "act2vec, trace2vec, log2vec, and model2vec: Representation Learning for Business Processes." *International Conference on Business Process Management*. Springer, Cham, 2018. |
| **Attri2Vec** | Zhang, Daokun, et al. "Attributed network embedding via subspace discovery." Data Mining and Knowledge Discovery 33.6 (2019): 1953-1980. |
| **Author2Vec** | Ganguly, Soumyajit, et al. "Author2vec: Learning author representations by combining content and link information." *Proceedings of the 25th International Conference Companion on World Wide Web*. International World Wide Web Conferences Steering Committee, 2016. |
| **Bib2Vec** | Yoneda, Takuma, et al. "Bib2vec: An Embedding-based Search System for Bibliographic Information." *arXiv preprint arXiv:1706.05122* (2017). |
| **Char2Vec** | Hussain, Syed-Amad A., et al. "Char2Vec: Learning the Semantic Embedding of Rare and Unseen Words in the Biomedical Literature." *AMIA*. 2018. |
| **Client2Vec** | Baldassini, Leonardo, and Jose Antonio Rodríguez Serrano. "client2vec: towards systematic baselines for banking applications." *arXiv preprint arXiv:1802.04198* (2018). |
| **Code2vec** | Kartchner, David, et al. "Code2vec: Embedding and clustering medical diagnosis data." *2017 IEEE International Conference on Healthcare Informatics (ICHI)*. IEEE, 2017. |
| **Code2vec** | Alon, Uri, et al. "code2vec: Learning distributed representations of code." *Proceedings of the ACM on Programming Languages* 3.POPL (2019): 1-29. |
| **Confusion2vec** | Shivakumar, Prashanth Gurunath, and Panayiotis Georgiou. "Confusion2Vec: towards enriching vector space word representations with representational ambiguities." *PeerJ Computer Science* 5 (2019): e195. |
| **Content2vec** | Nedelec, Thomas, Elena Smirnova, and Flavian Vasile. "Content2vec: Specializing joint representations of product images and text for the task of product recommendation." (2016). |
| **Crisis2vec** | Liu, Junhua, et al. "CrisisBERT: Robust Transformer for Crisis Classification and Contextual Crisis Embedding." *arXiv preprint arXiv:2005.06627* (2020). |
| **doc2vec** | Le, Quoc, and Tomas Mikolov. "Distributed representations of sentences and documents." *International conference on machine learning*. 2014. |
| **Domain2vec** | Deshmukh, Aniket Anand, Ankit Bansal, and Akash Rastogi. "Domain2Vec: Deep Domain Generalization." *arXiv preprint arXiv:1807.02919* (2018). |
| **dyngraph2vec** | Goyal, Palash, Sujit Rokka Chhetri, and Arquimedes Canedo. "dyngraph2vec: Capturing network dynamics using dynamic graph representation learning." *Knowledge-Based Systems* 187 (2020): 104816. |
| **dynnode2vec** | Mahdavi, Sedigheh, Shima Khoshraftar, and Aijun An. "dynnode2vec: Scalable dynamic network embedding." *2018 IEEE International Conference on Big Data (Big Data)*. IEEE, 2018. |
| **Edge2vec** | Gao, Zheng, et al. "edge2vec: Representation learning using edge semantics for biomedical knowledge discovery." *BMC bioinformatics* 20.1 (2019): 306. |
| **Event2vec** | Hong, Shenda, et al. "Event2vec: Learning representations of events on temporal sequences." *Asia-Pacific Web (APWeb) and Web-Age Information Management (WAIM) Joint Conference on Web and Big Data*. Springer, Cham, 2017. |
| **Event2vec** | Fu, Guoji, et al. "Representation Learning for Heterogeneous Information Networks via Embedding Events." *International Conference on Neural Information Processing*. Springer, Cham, 2019. |
| **Gene2vec** | Du, Jingcheng, et al. "Gene2vec: distributed representation of genes based on co-expression." *BMC genomics* 20.1 (2019): 82. |
| **graph2vec** | Narayanan, Annamalai, et al. "graph2vec: Learning distributed representations of graphs." *arXiv preprint arXiv:1707.05005* (2017). |
| **Hashtag2vec** | Liu, Jie, Zhicheng He, and Yalou Huang. "Hashtag2Vec: Learning Hashtag Representation with Relational Hierarchical Embedding Model." *IJCAI*. 2018. |
| **Inf2vec** | Feng, Shanshan, et al. "Inf2vec: Latent representation model for social influence embedding." 2018 IEEE 34th International Conference on Data Engineering (ICDE). IEEE, 2018. |
| **Inpatient2vec** | Wang, Ying, et al. "Inpatient2vec: Medical representation learning for inpatients." *2019 IEEE International Conference on Bioinformatics and Biomedicine (BIBM)*. IEEE, 2019. |
| **Inst2vec** | Ben-Nun, Tal, Alice Shoshana Jakobovits, and Torsten Hoefler. "Neural code comprehension: A learnable representation of code semantics." *Advances in Neural Information Processing Systems*. 2018. |

**Job2vec** Zhang, Denghui, et al. "Job2Vec: Job Title Benchmarking with Collective Multi-View Representation Learning." *Proceedings of the 28th ACM International Conference on Information and Knowledge Management*. 2019.

**lang2vec** Littell, Patrick, et al. "Uriel and lang2vec: Representing languages as typological, geographical, and phylogenetic vectors." *Proceedings of the 15th Conference of the European Chapter of the Association for Computational Linguistics: Volume 2, Short Papers*. 2017.

**lda2vec** Moody, Christopher E. "Mixing dirichlet topic models and word embeddings to make lda2vec." *arXiv preprint arXiv:1605.02019* (2016).

**med2vec** Choi, Edward, et al. "Multi-layer representation learning for medical concepts." *Proceedings of the 22nd ACM SIGKDD International Conference on Knowledge Discovery and Data Mining*. 2016.

**MEgo2vec** Zhang, Jing, et al. "MEgo2Vec: Embedding matched ego networks for user alignment across social networks." *Proceedings of the 27th ACM International Conference on Information and Knowledge Management*. 2018.

**Melody2vec** Hirai, Tatsunori, and Shun Sawada. "Melody2Vec: Distributed Representations of Melodic Phrases based on Melody Segmentation." *Journal of Information Processing* 27 (2019): 278-286.

**Metagraph2vec** Zhang, Daokun, et al. "Metagraph2vec: Complex semantic path augmented heterogeneous network embedding." Pacific-Asia conference on knowledge discovery and data mining. Springer, Cham, 2018.

**metapath2vec** Dong, Yuxiao, Nitesh V. Chawla, and Ananthram Swami. "metapath2vec: Scalable representation learning for heterogeneous networks." *Proceedings of the 23rd ACM SIGKDD international conference on knowledge discovery and data mining*. 2017.

**Mrnet-product2vec** Biswas, Arijit, Mukul Bhutani, and Subhajit Sanyal. "Mrnet-product2vec: A multi-task recurrent neural network for product embeddings." *Joint European Conference on Machine Learning and Knowledge Discovery in Databases*. Springer, Cham, 2017.

**News2vec** Ma, Ye, et al. "News2vec: News Network Embedding with Subnode Information." *Proceedings of the 2019 Conference on Empirical Methods in Natural Language Processing and the 9th International Joint Conference on Natural Language Processing (EMNLP-IJCNLP)*. 2019.

**node2vec** Grover, Aditya, and Jure Leskovec. "node2vec: Scalable feature learning for networks." *Proceedings of the 22nd ACM SIGKDD international conference*

| | |
|---|---|
| | *on Knowledge discovery and data mining*. 2016. |
| **Object2vec** | Bonner, Michael F., and Russell A. Epstein. "Object representations in the human brain reflect the co-occurrence statistics of vision and language." *bioRxiv* (2020). |
| **Onto2vec** | Smaili, Fatima Zohra, Xin Gao, and Robert Hoehndorf. "Onto2vec: joint vector-based representation of biological entities and their ontology-based annotations." *Bioinformatics* 34.13 (2018): i52-i60. |
| **OWL2Vec** | Holter, Ole Magnus, et al. "Embedding OWL ontologies with OWL2Vec." *CEUR Workshop Proceedings*. Vol. 2456. 2019. |
| **Paper2Vec** | Ganguly, Soumyajit, and Vikram Pudi. "Paper2vec: Combining graph and text information for scientific paper representation." *European Conference on Information Retrieval*. Springer, Cham, 2017. |
| **Path2vec** | Kutuzov, Andrey, et al. "Making Fast Graph-based Algorithms with Graph Metric Embeddings." *arXiv preprint arXiv:1906.07040* (2019). |
| **Patient2Vec** | Zhang, Jinghe, et al. "Patient2vec: A personalized interpretable deep representation of the longitudinal electronic health record." *IEEE Access* 6 (2018): 65333-65346. |
| **People2vec** | Kumar, Sumeet, and Kathleen M. Carley. "People2vec: Learning latent representations of users using their social-media activities." *International Conference on Social Computing, Behavioral-Cultural Modeling and Prediction and Behavior Representation in Modeling and Simulation*. Springer, Cham, 2018. |
| **Post2Vec** | Xu, Bowen, et al. "Post2Vec: Learning Distributed Representations of Stack Overflow Posts." IEEE Transactions on Software Engineering (2021). |
| **Rank2vec** | Zhou, Hui, et al. "Rank2vec: Learning node embeddings with local structure and global ranking." Expert Systems with Applications 136 (2019): 276-287. |
| **RDF2Vec** | Ristoski, Petar, et al. "RDF2Vec: RDF graph embeddings and their applications." *Semantic Web* 10.4 (2019): 721-752. |
| **Record2vec** | Sim, Adelene YL, and Andrew Borthwick. "Record2Vec: unsupervised representation learning for structured records." *2018 IEEE International Conference on Data Mining (ICDM)*. IEEE, 2018. |
| **Resource2Vec** | Coucheiro-Limeres, Alejandro, et al. "Resource2Vec: Linked Data distributed representations for term discovery in automatic speech recognition." Expert Systems with Applications 112 (2018): 301-320. |
| **sense2vec** | Trask, Andrew, Phil Michalak, and John Liu. "sense2vec-a fast and accurate method for word sense disambiguation in neural word embeddings." *arXiv preprint arXiv:1511.06388* (2015). |
| **Sent2vec** | Pagliardini, Matteo, Prakhar Gupta, and Martin Jaggi. "Unsupervised learning of sentence embeddings using compositional n-gram features." *arXiv preprint arXiv:1703.02507* (2017). |
| **Space2vec** | Mai, Gengchen, et al. "Multi-scale representation learning for spatial feature distributions using grid cells." *arXiv preprint arXiv:2003.00824* (2020). |
| **struc2vec** | Ribeiro, Leonardo FR, Pedro HP Saverese, and Daniel R. Figueiredo. "struc2vec: Learning node representations from structural identity." *Proceedings of the 23rd ACM SIGKDD International Conference on Knowledge Discovery and Data Mining*. 2017. |
| **Subgraph2vec** | Narayanan, Annamalai, et al. "subgraph2vec: Learning distributed representations of rooted sub-graphs from large graphs." arXiv preprint arXiv:1606.08928 (2016). |
| **Sub2vec** | Adhikari, Bijaya, et al. "Sub2vec: Feature learning for subgraphs." Pacific-Asia Conference on Knowledge Discovery and Data Mining. Springer, Cham, 2018. |
| **Time2vec** | Kazemi, Seyed Mehran, et al. "Time2Vec: Learning a Vector Representation of Time." *arXiv preprint arXiv:1907.05321* (2019). |
| **Topic2Vec** | Niu, Liqiang, et al. "Topic2Vec: learning distributed representations of topics." *2015 International conference on asian language processing (IALP)*. IEEE, 2015. |
| **Tweet2vec** | Vosoughi, Soroush, Prashanth Vijayaraghavan, and Deb Roy. "Tweet2vec: Learning tweet embeddings using character-level cnn-lstm encoder-decoder." *Proceedings of the 39th International ACM SIGIR conference on Research and Development in Information Retrieval*. 2016. |
| **URL2Vec** | Yuan, Huaping, et al. "URL2Vec: URL Modeling with Character Embeddings for Fast and Accurate Phishing Website Detection." *2018 IEEE Intl Conf on Parallel & Distributed Processing with Applications, Ubiquitous Computing & Communications, Big Data & Cloud Computing, Social Computing & Networking, Sustainable Computing & Communications (ISPA/IUCC/BDCloud/SocialCom/SustainCom)*. IEEE, 2018. |
| **user2Vec** | Hallac, Ibrahim R., et al. "user2Vec: Social Media User Representation Based on Distributed Document Embeddings." *2019 International Artificial* |
| | *Intelligence and Data Processing Symposium (IDAP)*. IEEE, 2019. |
| **Video2vec** | Hu, Sheng-Hung, Yikang Li, and Baoxin Li. "Video2vec: Learning semantic spatio-temporal embeddings for video representation." *2016 23rd International Conference on Pattern Recognition (ICPR)*. IEEE, 2016. |
| **W-Metagraph2Vec** | Pham, Phu, and Phuc Do. "W-Metagraph2Vec: a novel approval of enriched schematic topic-driven heterogeneous information network embedding." International Journal of Machine Learning and Cybernetics (2020): 1-20. |
| **Wave2Vec** | Yuan, Ye, et al. "Wave2vec: Deep representation learning for clinical temporal data." *Neurocomputing* 324 (2019): 31-42. |